\documentclass[10pt,journal,compsoc]{IEEEtran}
\ifCLASSOPTIONcompsoc
  \usepackage[nocompress]{cite}
\else
  \usepackage{cite}
\fi
\usepackage{graphicx}
\usepackage{epstopdf}
\graphicspath{ {./figs/} }

\ifCLASSINFOpdf
\else
\fi
\usepackage{amsmath}
\usepackage{algorithmic}

\usepackage{algorithm}

\ifCLASSOPTIONcompsoc
  \usepackage[caption=false,font=footnotesize,labelfont=sf,textfont=sf]{subfig}
\else
  \usepackage[caption=false,font=footnotesize]{subfig}
\fi
\usepackage{tikz,tikz-3dplot}
\usetikzlibrary{arrows}
\usetikzlibrary{calc,fit}

\usepackage{enumitem}

\usepackage{amsthm}

\newtheorem{assumption}{Assumption}
\newtheorem{theorem}{Theorem}

\newtheorem{lemma}{Lemma}

\theoremstyle{definition}
\newtheorem{definition}{Definition}
\theoremstyle{remark}

\theoremstyle{remark}

\theoremstyle{definition}

\usepackage{amssymb} 

\newcommand{\myparagraph}[1]{\smallskip\noindent\textbf{#1.}}
\newcommand{\RR}{I\!\!R} 

\def\Re{\mathbb{R}}
\def\Sp{\mathbb{S}}

\DeclareMathOperator*{\argmin}{arg\,min}
\DeclareMathOperator*{\argmax}{arg\,max}

\def\0{\boldsymbol{0}}
\def\x{\boldsymbol{x}}
\def\e{\boldsymbol{e}}
\def\c{\boldsymbol{c}}
\def\r{\boldsymbol{r}}
\def\v{\boldsymbol{v}}
\def\w{\boldsymbol{w}}
\def\A{\mathbf{A}}
\def\W{\mathbf{W}}
\def\cX{\mathcal{X}}
\def\cC{\mathcal{C}}

\def\cS{\mathcal{S}}
\def\cV{\mathcal{V}}
\def\cK{\mathcal{K}}
\def\transpose{\top}

\def\st{~~\mathrm{s.t.}~~}

\def\spann{\text{span}}

\begin{document}
%
\title{Self-Representation Based Unsupervised Exemplar Selection in a Union of Subspaces}
%
%
%
%

\author{Chong~You,~
	    Chi~Li,~
        Daniel~P.~Robinson,~
        and~Ren\'e~Vidal,~\IEEEmembership{Fellow,~IEEE}
\IEEEcompsocitemizethanks{
	\IEEEcompsocthanksitem C. You is with the Department of Electrical Engineering and
	Computer Science, University of California, Berkeley, USA. Most of the work was done while C. You was at The Johns Hopkins University, USA.\protect\\
	E-mail: cyou@berkeley.edu.
	\IEEEcompsocthanksitem C. Li is a machine learning scientist in Apple Inc, Cupertino, USA.\protect\\
	E-mail: chi\_li@jhu.edu.
	\IEEEcompsocthanksitem D. Robinson is with the Department of Industrial and Systems Engineering, Lehigh University, USA.\protect\\
	E-mail: daniel.p.robinson@gmail.edu.
	\IEEEcompsocthanksitem R. Vidal is with the Department of
	Biomedical Engineering, The Johns Hopkins University, USA.\protect\\
	E-mail: rvidal@cis.jhu.edu.
}
}

\IEEEtitleabstractindextext{%
\begin{abstract}
Finding a small set of representatives from an unlabeled dataset is a core problem in a broad range of applications such as dataset summarization and information extraction. Classical exemplar selection methods such as $k$-medoids work under the assumption that the data points are close to a few cluster centroids, and cannot handle the case where data lie close to a union of subspaces.
This paper proposes a new exemplar selection model that searches for a subset that best reconstructs all data points as measured by the $\ell_1$ norm of the representation coefficients. Geometrically, this subset best covers all the data points as measured by the Minkowski functional of the subset. To solve our model efficiently, we introduce a farthest first search algorithm that iteratively selects the worst represented point as an exemplar. When the dataset is drawn from a union of independent subspaces, our method is able to select sufficiently many representatives from each subspace. We further develop an exemplar based subspace clustering method that is robust to imbalanced data and efficient for large scale data. Moreover, we show that a classifier trained on the selected exemplars (when they are labeled) can correctly classify the rest of the data points.
\end{abstract}

}

\maketitle

\IEEEdisplaynontitleabstractindextext

%
\IEEEpeerreviewmaketitle

\IEEEraisesectionheading{\section{Introduction}\label{sec:introduction}}

%
%
%
%

\begin{figure}[!h]
	\centering
	\centering
	\includegraphics[width=0.92\linewidth]{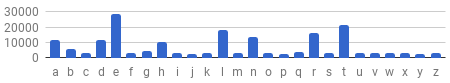}
	\\
	\vspace{1em}
	\includegraphics[width=0.92\linewidth,trim={0cm 13cm 6cm 0cm},clip]{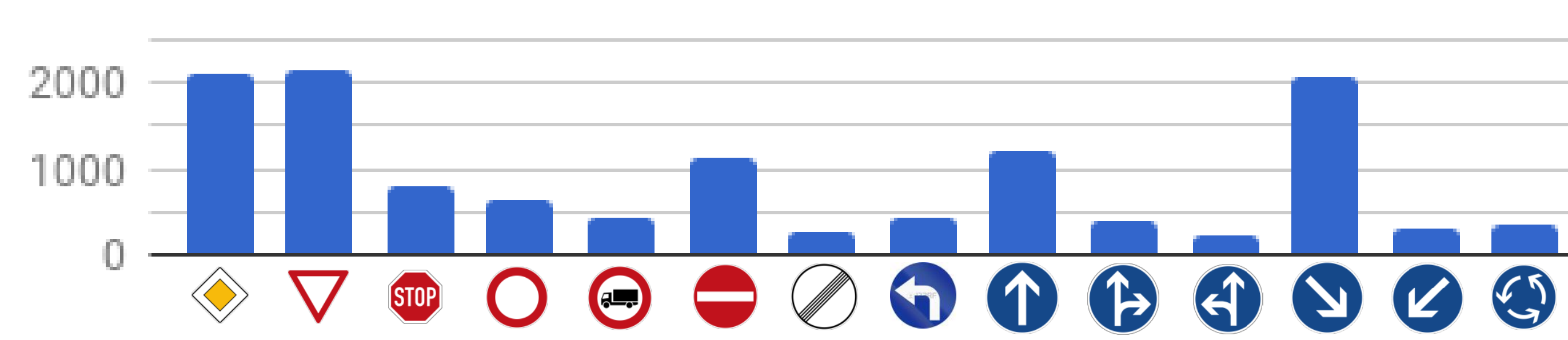}
	\caption{\label{fig:frequency}Number of points in each  class associated with the EMNIST handwritten letters (top) and the GTSRB (bottom) street sign databases. 
	}
\end{figure}

\IEEEPARstart{T}{he} availability of large annotated datasets in computer vision, such as ImageNet, has led to many recent breakthroughs in object detection and classification using supervised learning techniques such as deep learning. 
However, as data sizes continue to grow, it has become difficult to annotate the data for training fully supervised algorithms.
As a consequence, the development of unsupervised learning techniques that can learn from \emph{unlabeled} datasets has become extremely important. 
In addition to the challenge introduced by the sheer volume of data, the number of data samples in unlabeled datasets usually varies widely for different classes.
{
For example, a street sign database collected from street view images may contain drastically different numbers of instances for different types of signs since not all of them are used on streets with the same frequency; a handwritten letter database may be highly imbalanced as the frequency of different letters in English text varies significantly (see Figure~\ref{fig:frequency}). 
}
An imbalanced data distribution is known to compromise performance of canonical supervised \cite{Akbani:ECML04} and unsupervised \cite{Xiong:TSMC09} learning techniques.

We exploit the idea of exemplar selection to address the challenge of learning from an unlabeled dataset.
Exemplar selection refers to the problem of selecting a set of data representatives or exemplars from the data.
It has been a particularly useful approach for scaling up existing data clustering algorithms so that they can handle large datasets more efficiently \cite{Chen:AAAI11}.
Finding an exemplar set that is informative of the entire data is often the key challenge for the success of such approaches.
Particularly, when the data is drawn from several different groups, it is crucial that an algorithm selects enough samples from each of the groups without prior knowledge of which points belong to which groups.
This can be especially difficult when the data is imbalanced, as it is more likely to select data from over-represented groups than from under-represented groups.

%

Exemplar selection is also useful when one has limited resources so that only a small subset of data can be labeled.
In such cases, exemplar selection can determine the subset to be manually labeled, and then used to train a model to infer labels for the remaining data \cite{Wei:ICML15}.
The ability to correctly classify as many of the unlabeled data points as possible depends critically on the quality of the selected exemplars.

Some of the most popular methods for exemplar selection include $k$-centers and $k$-medoids, which search for the set of centers and medoids that best fit the data under the assumption that data points concentrate around a few discrete points. 
However, certain high-dimensional image and video data is distributed among certain low-dimensional subspaces \cite{Tomasi:IJCV92,Basri:PAMI03}, and the discrete center based methods become ineffective.
\emph{In this paper, we consider exemplar selection under a model where the data points lie close to a collection of unknown low-dimensional subspaces.} 
One line of work that can address such problem is based on the assumption that each data point can be expressed by a few data representatives with small reconstruction residual. 
This includes the simultaneous sparse representation \cite{Tropp:SP06I} and dictionary selection \cite{Cevher:JSTSP11,Das:arxiv11}, which use greedy algorithms to solve their respective optimization problems, and group sparse representative selection \cite{Tropp:SP06,Cong:CVPR11,Elhamifar:CVPR12,Meng:CVPR16,Wang:PR17,Cong:TMM}, which uses a convex optimization approach based on group sparsity. 
In particular, the analysis in \cite{Elhamifar:CVPR12} shows that when data come from a union of subspaces, their method is able to select a few representatives from each of the subspaces.
However, methods in this category cannot effectively handle large-scale data as they have quadratic complexity in the number of points. 
Moreover, the convex optimization based methods such as that in \cite{Elhamifar:CVPR12} are not flexible in selecting a desired number of representatives since the size of the subset cannot be directly controlled by adjusting an algorithm parameter.

\subsection{Paper contributions}
We present a data self-representation based exemplar selection algorithm for learning from large scale and imbalanced data in an unsupervised manner.
Our method is based on the \emph{self-expressiveness} property of data in a union of subspaces \cite{Elhamifar:CVPR09}, which states that each data point in a union of subspaces can be written as a linear combination of other points from its own subspace. 
That is, given data $\cX=\{\x_1, \cdots, \x_N\}\subseteq \RR^D$, there exists $\{c_{ij}\}$ such that $\x_j = \sum_{i\ne j}c_{ij} \x_i$ and $c_{ij}$ is nonzero only if $\x_i$ and $\x_j$ are from the same subspace. 
Such representations $\{c_{ij}\}$ are called \emph{subspace-preserving}.
In particular, if the subspace dimensions are small, then the representations can be taken to be sparse. 
Based on this observation, 
\cite{Elhamifar:CVPR09} proposes the Sparse Subspace Clustering (SSC) method, which computes for each $\x_j\in\cX$
the vector $\c_j=[c_{1j}, \cdots, c_{Nj}]^\top$ as a solution to the sparse optimization problem
\begin{equation}\label{eq:ssc}
\min_{\c \in \Re^N} \ \|\c\|_1 + \tfrac{\lambda}{2}  \|\x_j - \sum_{i\ne j} c_{i}\x_i\|_2^2,
\end{equation}
where $\lambda>0$.
In \cite{Elhamifar:CVPR09}, the solution to \eqref{eq:ssc} is used to define an affinity between any pair of points $\x_i$ and $\x_j$ as $|c_{ij}|+|c_{ji}|$, and then spectral clustering is applied to generate a segmentation of the data points into their respective subspaces. 
Existing theoretical results show that, under certain assumptions on the data, the solution to \eqref{eq:ssc} is subspace-preserving \cite{Elhamifar:TPAMI13,Soltanolkotabi:AS12,You:ICML15,Wang:ICML15,Wang:JMLR16,You:CVPR17,Li:JSTSP18,Robinson:arxiv19,You:ICCV19}, thus justifying the correctness of the affinity produced by SSC. 

While the nonzero entries for each $\c_j$ determine a subset of $\cX$ that can represent $\x_j$ with the minimum $\ell_1$-norm on the coefficients, the union of the representations over all $\{\c_j\}$ often uses the entire dataset $\cX$. In this paper, we propose to find a small subset $\cX_0 \subseteq \cX$, which we call \emph{exemplars}, such that solutions $\c_j$ to the problem
\begin{equation}
\min_{\c \in \Re^N} \|\c\|_1 + \tfrac{\lambda}{2} \|\x_j - \sum_{i: \x_i \in \cX_0} c_{i}\x_i \|_2^2
\label{eq:esc}
\end{equation}
are also subspace-preserving. Since $\cX_0$ is a small subset of $\cX$, solving the optimization problem~\eqref{eq:esc} is much cheaper computationally compared to~\eqref{eq:ssc}. Computing an appropriate $\cX_0$ through an exhaustive search would be computationally impractical.
To address this issue, we present an efficient algorithm (an exemplar selection algorithm) that iteratively selects the worst represented point from the data $\cX$ to form $\cX_0$.
Our exemplar selection procedure is then used to design an exemplar-based subspace clustering approach (assuming that the exemplars are unlabeled) \cite{You:ECCV18} and an exemplar-based classification approach (assuming that the exemplars are labeled) by using the representative power of the selected exemplars.
In summary, our work makes the following contributions compared to the state of the art:

\begin{figure}[t]
	\centering
	\includegraphics[scale=0.8]{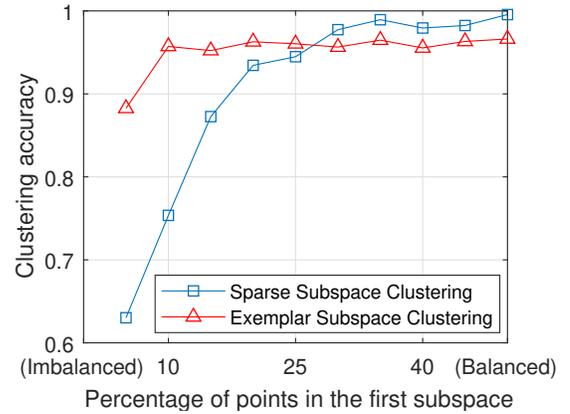}
	\caption{\label{fig:overview_imbalanced} Subspace clustering on imbalanced data. Two  subspaces of dimension three are generated uniformly at random in ambient space of dimension five. Then, $x$ and $100-x$ points are sampled uniformly at random from the two subspaces, respectively, where $x$ is varied in the x-axis. The clustering accuracy of SSC decreases dramatically as the dataset becomes imbalanced. The exemplar based subspace clustering (see Algorithm~\ref{alg:esc}) is more robust to imbalanced data distribution.  }
\end{figure}

\begin{itemize}[leftmargin=*,topsep=0.3em,noitemsep]
	\item We present a geometric interpretation of our exemplar selection algorithm as one of finding a subset of the data that \emph{best covers} the entire dataset as measured by the Minkowski functional of the subset.
	When the data lies in a union of independent subspaces, we prove that our method selects sufficiently many representative data points (exemplars) from each subspace, even when the dataset is imbalanced.
	Unlike prior methods such as \cite{Elhamifar:CVPR12}, our method has linear execution time and memory complexity in the number of data points for each iteration, and can be terminated when the desired number of exemplars have been selected. 
	
	\item We show that the exemplars in $\cX_0$ selected by our method can be used for subspace clustering by first computing the representations for each data point with respect to the exemplars as in \eqref{eq:esc}, second constructing a $k$-nearest neighbor graph of the representation vectors, and third applying spectral clustering. 
	Compared to SSC, the exemplar-based subspace clustering method is empirically less sensitive to imbalanced data and more efficient on large-scale datasets (see Figure~\ref{fig:overview_imbalanced}).
	Experimental results on the large-scale and label-imbalanced handwritten letter dataset EMNIST and street sign dataset GTSRB show that our method outperforms state-of-the-art algorithms in terms of both clustering performance and running time.
	
	\item We show that a classifier trained on the exemplars selected by our model (assuming that the labels of the exemplars are provided) is able to correctly classify the rest of the data points.
	We demonstrate through experiments on the Extended Yale B face database that exemplars selected by our method produce higher classification accuracy when compared to several popular exemplar selection methods.
\end{itemize}

{
We remark that a conference version of the paper appeared in the proceedings of European Conference on Computer Vision (ECCV) in 2018 \cite{You:ECCV18}. 
In comparison to the conference version, which focuses on the problem of subspace clustering on imbalanced data, the current paper addresses the problem of exemplar selection, which has a broader range of applications that include data summarization, clustering and classification tasks. 
With additional technical results and experimental evaluation, the current paper provides a more
comprehensive study of the subject.
}

\subsection{Related work}
\label{sec:related-work}

\myparagraph{Exemplar selection} 
Two of the most popular methods for exemplar selection are $k$-centers and $k$-medoids. The $k$-centers problem is a data clustering problem studied in theoretical computer science and operations research. Given a set $\cX$ and an integer $k$, the goal is to find a set of centers $\cX_0 \subseteq \cX$ with $|\cX_0| \le k$ that minimizes the quantity $\max_{\x \in \cX} d^2(\x, \cX_0)$, where $d^2(\x, \cX_0):=\min_{\v \in \cX_0} \|\x - \v\|_2^2$ is the squared distance of $\x$ to the closest point in $\cX_0$. A partition of $\cX$ is given by the closest center to which each point $\x \in \cX$ belongs. 
The $k$-medoids is a variant of $k$-centers that minimizes the sum of the squared distances, i.e., minimizes $\sum_{\x \in \cX} d^2(\x, \cX_0)$ instead of the maximum distance. However, both $k$-centers and $k$-medoids model data as concentrating around several cluster centers, and do not generally apply to data lying in a union of subspaces. 

In general, selecting a representative subset of the entire data has been studied in a wide range of contexts such as 
Determinantal Point Processes ~\cite{Borodin:09arxiv,Gillenwater:NIPS14,Kulesza:ICML11},
Prototype Selection~\cite{Garcia:TPAMI12,Liu:KBS17},
Rank Revealing QR~\cite{Chan:LAA87},
Column Subset Selection (CSS)~\cite{Boutsidis:SODA09,Altschuler:ICML16,Joneidi:CVPR2019,Joneidi:CVPR20}, separable Nonnegative Matrix Factorization (NMF)~\cite{Gillis:TPAMI13,Arora:ACM12,Kumar:ICML13}, and so on~\cite{Elhamifar:NIPS12}. 
In particular, both CSS and separable NMF can be interpreted as finding exemplars such that each data point can be expressed as a linear combination of such exemplars.
However, these methods do not impose sparsity on the representation coefficients, and therefore cannot be used to select good representatives from data that is drawn from a union of low-dimensional subspaces.

\myparagraph{Subspace clustering on imbalanced and large scale data} 
Subspace clustering aims to cluster data points drawn from a union of subspaces into their respective subspaces.
Recently, self-expressiveness based subspace clustering methods such as SSC and its variances \cite{Liu:ICML10,Lu:ECCV12,Dyer:JMLR13,Yang:ECCV16,Ji:NIPS17,Xin:TSP18,Chen:CVPR20} have achieved great success for many computer vision tasks such as face clustering, handwritten digit clustering, and so on. 
Nonetheless, previous experimental evaluations focused primarily on balanced datasets, i.e. datasets with approximately the same number of samples from each cluster. In practice, datasets are often imbalanced and such skewed data distributions can significantly compromise the clustering performance of SSC.
There has been no study of this issue in the literature to the best of our knowledge.

Another issue with many self-expressive based subspace clustering methods is that they are limited to small or medium scale datasets \cite{You:Asilomar16}.
Several works addressed the scalability issue by computing a dictionary with number of atoms much smaller than the total number of data points in $\cX$, and expressing each data point in $\cX$ as a linear combination of the atoms in the dictionary (the dictionary is usually not a subset of $\cX$). 
In particular, \cite{Adler:TNNLS15} shows that if the atoms in the dictionary happen to lie in the same union of subspaces as the input data $\cX$, then this approach is guaranteed to be correct.
However, there is little evidence that such a condition is satisfied for real data as the atoms of the dictionary are not constrained to be a subset of $\cX$. Another recent work \cite{Traganitis:TSP17}, which uses data-independent random matrices as dictionaries, also suffers from this issue and lacks correctness guarantees.
More recently, several works \cite{Aldroubi:ACHA17,Aldroubi:arxiv17,Abdolali:arxiv18} use exemplar selection to form the dictionary for subspace clustering, but they lack theoretical justification that their selected exemplars represent the subspaces.

\section{Self-Representation based Unsupervised Exemplar Selection}
\label{sec:ESC}

In this section, we present our self-representation based method for exemplar selection from an unlabeled dataset $\cX=\{\x_1, \cdots, \x_N\}$, which are assumed to have unit $\ell_2$ norm.\footnote{This is not a strong assumption as one can always normalize the data points as a preprocessing step for any given dataset. } 
We first formulate the model for selecting a subset $\cX_0$ of exemplars from $\cX$ in Section~\ref{sec:ESC_model} as minimizing a self-representation cost. 
Since the model is a combinatorial optimization problem, we present an efficient algorithm for solving it approximately in Section~\ref{sec:FFS}. 

\subsection{A self-representation cost for exemplar selection}
\label{sec:ESC_model}


In our exemplar selection model, the goal is to find a small subset $\cX_0 \subseteq \cX$ that can linearly represent all data points in $\cX$. In particular, the set $\cX_0$ should contain exemplars from each subspace such that the solution to \eqref{eq:esc} for each data point $\x_j \in \cX$ is subspace-preserving.
Next, we define a cost function based on the optimization problem in \eqref{eq:esc} and then present our exemplar selection model.

\begin{definition}[Self-representation cost function]
	\label{def:representation-cost}
	Given $\cX = \{\x_1, \cdots, \x_N\}$ $\subseteq \Re^D$, we define the self-representation cost function $F_\lambda: 2^{\cX} \to \Re$ as
	\begin{equation}\label{eq:representation-cost}
	F_\lambda(\cX_0):= \sup_{\x_j \in \cX} f_\lambda(\x_j, \cX_0) 
	\end{equation}
	where
	\begin{equation}\label{eq:representation-f}
	f_\lambda(\x_j, \cX_0) := \min_{\c\in \Re^N} \|\c\|_1 + \tfrac{\lambda}{2} \|\x_j - \sum_{i: \x_i \in \cX_0} c_i \x_i\|_2^2
	\end{equation}
	and $\lambda \in (1, \infty)$ is a parameter. By convention, we define $f_\lambda(\x_j, \emptyset) = \frac{\lambda}{2}$ for all $\x_j\in\cX$, where $\emptyset$ is the empty set.
\end{definition}

The quantity $f_\lambda(\x_j, \cX_0)$ is a measure of how well the data point $\x_j\in \cX$ is represented by the subset $\cX_0$.
The function $f_\lambda(\x_j, \cX_0)$ has the following properties.
\begin{lemma}\label{thm:f-monotone}
	For each $j\in\{1,\dots, N\}$, the function $f_\lambda(\x_j, \cdot)$ is monotone with respect to the partial order defined by set inclusion, i.e., $f_\lambda(\x_j, \cX_0') \ge f_\lambda(\x_j, \cX_0'')$ for any $\emptyset \subseteq \cX_0' \subseteq  \cX_0'' \subseteq \cX$. 
\end{lemma}

\begin{proof}
	Let $j\in\{1,\dots N\}$.  Then, let us define $\c'$ as 
	\begin{equation*} 
	\c' = [c_1', \cdots, c_N']^\transpose \in \argmin_{\c\in \Re^N} \|\c\|_1 + \tfrac{\lambda}{2}\|\x_j - \sum_{i: \x_i \in \cX_0'} c_{i} \x_i\|_2^2.
	\end{equation*}
	It follows from the optimality conditions 
	that $c_i' = 0$ for all $i$
	such that $\x_i \notin \cX_0'$. Combining this with  $\cX_0' \subseteq  \cX_0''$ yields
	\begin{equation*}
	\begin{split}
		f_\lambda(\x_j, \cX_0') 
		&= \|\c'\|_1 + \tfrac{\lambda}{2}\|\x_j - \sum_{i: \x_i \in \cX_0'} c_{i}' \x_i\|_2^2 \\
		&= \|\c'\|_1 + \tfrac{\lambda}{2}\|\x_j - \sum_{i: \x_i \in \cX_0''} c_{i}' \x_i\|_2^2 \\
		&\geq \min_{\c\in \Re^N} \|\c\|_1 + \tfrac{\lambda}{2}\|\x_j - \sum_{i: \x_i \in \cX_0''} c_i \x_i\|_2^2 \\
		&= f_\lambda(\x_j,\cX_0^{''}),
	\end{split}
	\end{equation*}
	which is the desired result.
\end{proof}

\begin{lemma}\label{thm:f-range}
	For each $j\in\{1,\dots,N\}$ the following hold: (i) for every $\cX_0\in2^\cX$ the inclusion  $f_\lambda(\x_j, \cX_0) \in [1 - \frac{1}{2\lambda}, \frac{\lambda}{2}]$ holds; (ii) $f_\lambda(\x_j, \emptyset) = \lambda/2$; and (iii) $f_\lambda(\x_j, \cX_0) = 1 - \tfrac{1}{2\lambda}$ if and only if at least one of $\x_j$ or $-\x_j$ is in $\cX_0$.
\end{lemma}
\begin{proof}
First observe that if $\cX_0 = \emptyset$, then it follows from Definition \ref{def:representation-cost} that $f_\lambda(\x_j, \emptyset) = \lambda/2$. Second, consider the case $\cX_0 = \cX$.  In this case, define $\bar{\c} = [\bar{c}_{1}, \cdots, \bar{c}_{N}]$ to be the one-hot vector with $j$-th entry $\bar{c}_{j} = 1 - \frac{1}{\lambda}$ and all other entries zero. One can then verify that 
$\|\bar{\c}\|_1 +\frac{\lambda}{2}\|\x_j - \sum_{i=1}^N \bar{c}_{i}\x_i\|_2^2 = 1 - \frac{1}{2\lambda}$ {(by recalling the assumption that $\|\x_j\|_2=1$)}. Combining these two cases with Lemma~\ref{thm:f-monotone} establishes that parts (i) and (ii) hold.

For the ``if'' direction of part (iii), let either $\x_j \in \cX_0$ or $-\x_j \in \cX_0$.  
Define $\bar{\c} = [\bar{c}_{1}, \cdots, \bar{c}_{N}]$ as a one-hot vector with $j$-th entry 
$\bar{c}_{j} = 1 - \frac{1}{\lambda}$ if $\x_j \in \cX_0$, and $\bar{c}_{j} = -(1 - \frac{1}{\lambda})$ if $-\x_j \in \cX_0$; in either case all other entries are set to zero. One can then verify that $\|\bar{\c}\|_1 +\frac{\lambda}{2}\|\x_j - \sum_{i:\x_i \in \cX_0} \bar{c}_{i}\x_i\|_2^2 = 1 - \frac{1}{2\lambda}$, which completes the proof for this direction.
		
To prove the ``only if'' direction, suppose that  $f_\lambda(\x_j, \cX_0)=1-\frac{1}{2\lambda}$. 
Let us define 
\[
\c^* \in \arg\min_{\c} \|\c\|_1 + \tfrac{\lambda}{2}\|\x_j - \sum_{i: \x_i \in \cX_0} c_{i} \x_i\|_2^2
\]
and $\e^* = \x_j - \sum_{i: \x_i \in \cX_0} c_{i}^* \x_i$. 
From the optimality conditions, it follows that  $c^*_{i} = 0$ for all $i \in \{1, \cdots, N\}$ such that $\x_i \notin \cX_0$. 
Using this fact, the assumption that the data is normalized, and basic properties of norms, we have
\begin{equation}
\begin{split}
		1 &= \|\x_j\|_2 = \|\e^* + \sum_{i: \x_i \in \cX_0} c_{i}^* \x_i\|_2 \\
		&\le \|\e^*\|_2 + \|\sum_{i: \x_i \in \cX_0} c_{i}^* \x_i\|_2\\
		&\le \|\e^*\|_2 + \sum_{i: \x_i \in \cX_0} (|c_{i}^*| \|\x_i\|_2)
		= \|\e^*\|_2 + \|\c^*\|_1.
		\label{eq:prf-claim-ec-0}  
\end{split}           
\end{equation}
From \eqref{eq:prf-claim-ec-0}, $f_\lambda(\x_j, \cX_0)=1-\frac{1}{2\lambda}$ and definition of $\c^*$,  we have
\begin{align}
1 - \tfrac{1}{2\lambda}
&= f_\lambda(\x_j, \cX_0) \label{eq:prf-claim-lowerb-0}\\ 
&= \|\c^*\|_1 + \tfrac{\lambda}{2}\|\e^*\|_2^2  
\ge 1 - \|\e^*\|_2 + \tfrac{\lambda}{2}\|\e^*\|_2^2
\ge 1 - \tfrac{1}{2\lambda},  \nonumber
\end{align}
where the last inequality follows by computing the minimum value of $1 - \|\e^*\|_2 + \tfrac{\lambda}{2}\|\e^*\|_2^2$. 		
It follows that equality is achieved for all inequalities in \eqref{eq:prf-claim-lowerb-0}.
By requiring equality for the second and first inequalities in \eqref{eq:prf-claim-lowerb-0}, we get respectively,
\begin{equation}\label{eq:prf-opt-e}
  \|\e^*\|_2 = \tfrac{1}{\lambda}
  \ \ \text{and} \ \ 
  \|\c^*\|_1 = 1 - \tfrac{1}{\lambda}.
\end{equation} 
Since~\eqref{eq:prf-opt-e} implies $\|\e^*\|_2 + \|\c^*\|_1 = 1$, we can conclude that all of the inequalities in~\eqref{eq:prf-claim-ec-0} must actually be equalities.  Using this fact and~\eqref{eq:prf-claim-ec-0} we have that 
\begin{equation}\label{eq:prf-opt-comb}
\|\sum_{i: \x_i \in \cX_0} c_{i}^* \x_i\|_2
= 1 - \|\e^*\|_2 
= 1 - \tfrac{1}{\lambda}.
\end{equation}	
Define $\mu_0:=\max_{i: \x_i \in \cX_0} |\langle \x_j, \x_i \rangle|$.  From definition of $\e^*$, \eqref{eq:prf-opt-e}, the fact that the data is normalized, and~\eqref{eq:prf-opt-comb}, we have
\begin{multline}
\tfrac{1}{\lambda^2} = \|\e^*\|_2^2 = \|\x_j - \sum_{i: \x_i \in \cX_0} c_{i}^* \x_i\|_2^2 \\
= 1 - 2 \langle \x_j, \sum_{i: \x_i \in \cX_0} c_{i}^* \x_i \rangle + (1-\tfrac{1}{\lambda})^2.
\label{eq:prf-claim-contradict-mu-pre}
\end{multline}
For the second term on the right hand side of \eqref{eq:prf-claim-contradict-mu-pre}, we may use 
the fact that the data is normalized, definition of $\mu_0$, and~\eqref{eq:prf-opt-e} to conclude that
\begin{equation*}
\langle \x_j, \!\sum_{i: \x_i \in \cX_0} \!\!c_{i}^* \x_i \rangle 
= \!\sum_{i: \x_i \in \cX_0} \!c_{i}^* \langle \x_j, \x_i \rangle 
\le \mu_0 \|\c^*\|_1
= \mu_0(1 - \tfrac{1}{\lambda}).
\end{equation*}
Plugging this into \eqref{eq:prf-claim-contradict-mu-pre} yields
\begin{equation}
\tfrac{1}{\lambda^2} \ge 1 - 2 \mu_0  (1 - \tfrac{1}{\lambda}) + (1-\tfrac{1}{\lambda})^2,
\end{equation}
which after simplification shows that
\begin{equation}
0 
\ge -2 \mu_0 (1-\tfrac{1}{\lambda}) + 2(1-\tfrac{1}{\lambda})
= 2 (1-\tfrac{1}{\lambda}) (1-\mu_0).
\label{eq:prf-claim-contradict-mu}
\end{equation}
Recall that $\lambda \in (1, \infty)$ (see Definition \ref{def:representation-cost}). Therefore, from \eqref{eq:prf-claim-contradict-mu} we see that $\mu_0=\max_{i: \x_i \in \cX_0} |\langle \x_j, \x_i \rangle| \ge 1$. Since both $\x_j$ and $\x_i$ have unit $\ell_2$ norm, we conclude that $\mu_0 = 1$, i.e., that either $\x_j$ or $-\x_j$ must be in $\cX_0$, as desired.
\end{proof}


Observe that if $\cX_0$ contains enough exemplars from the subspace containing $\x_j$ and a solution $\c^*$ to the optimization problem in \eqref{eq:representation-f} is subspace-preserving, then it is expected that $\c^*$ will be sparse and that the residual $\x_j - \sum_{i: \x_i \in \cX_0} \x_i c_i^*$ will be close to zero. 
This suggests that we should select the subset $\cX_0$ such that the value $f_\lambda(\x_j, \cX_0)$ is small for all $j$.
As the value $F_\lambda(\cX_0)$ is achieved by the data point $\x_j$ that has the largest value $f(\x_j, \cX_0)$,
we propose to perform exemplar selection by searching for a subset $\cX_0^* \subseteq \cX$ that minimizes the self-representation cost function, i.e.,
\begin{equation}\label{eq:exemplar-objective}
\cX_0^* = \argmin_{|\cX_0| \le k} F_\lambda(\cX_0),
\end{equation}
\noindent
where $k\in \mathbb{Z}$ is the target number of exemplars.
The objective function $F_\lambda(\cdot)$ in \eqref{eq:exemplar-objective} is monotone as shown next.
\begin{lemma}\label{thm:exemplar-monotone}
	If $\emptyset \subseteq \cX_0'\subseteq \cX_0'' \subseteq \cX$, then $F_\lambda(\cX_0') \ge F_\lambda(\cX_0'')$. 
\end{lemma}

\begin{proof}
	Let us define
	$$
	\x' \in \arg\sup_{\x_j \in \cX} f_\lambda(\x_j, \cX_0')
	\ \ \text{and} \ \ 
	\x'' \in \arg\sup_{\x_j \in \cX} f_\lambda(\x_j, \cX_0'').
	$$
	It follows from these definitions and Lemma~\ref{thm:f-monotone} that 
	\begin{align*}
	F_\lambda(\cX_0') &= f_\lambda(\x', \cX_0') \\
	&\ge f_\lambda(\x'', \cX_0') 
	\ge f_\lambda(\x'', \cX_0'') = F_\lambda(\cX_0''),
	\end{align*}
	which completes the proof.
\end{proof}


\subsection{A Farthest First Search (FFS) algorithm}
\label{sec:FFS}

Solving the optimization problem \eqref{eq:exemplar-objective} is NP-hard in general as it requires evaluating $F_\lambda(\cX_0)$ for each subset $\cX_0$ of size at most $k$. 
In Algorithm~\ref{alg:ffs} below, we present a greedy algorithm for efficiently computing an approximate solution to \eqref{eq:exemplar-objective}. 
The algorithm progressively grows a candidate subset $\cX_0$ (initialized as the empty set) until it reaches the desired size $k$. During each iteration $i$, step~\ref{step:update} of the algorithm selects the point $\x_j \in \cX$ that is worst represented by the current subset $\cX_0^{(i)}$ as measured by $f_\lambda(\x_j, \cX_0^{(i)})$. 
It was shown in Lemma~\ref{thm:f-range} that $f_\lambda(\x_j, \cX_0^{(i)}) = 1 - \frac{1}{2\lambda}$ if $\x_j \in \cX_0^{(i)}$, and $f_\lambda(\x_j, \cX_0^{(i)}) > 1 - \frac{1}{2\lambda}$ if $\x_j \notin \cX_0^{(i)}$ and $-\x_j \notin \cX_0^{(i)}$.  Thus, during each iteration an element not from $\cX_0^{(i)}$ is added to $\cX_0^{(i)}$ to form $\cX_0^{(i+1)}$ when $N$ is sufficiently large.
When the algorithm terminates, the output $\cX_0^{(k)}$ contains exactly $k$ distinct exemplars from $\cX$. 



We also note that the FFS algorithm can be viewed as an extension of the farthest first traversal algorithm (see, e.g.~\cite{Williamson:Cambridge2011}), which is an approximation algorithm for the $k$-centers problem discussed in Section~\ref{sec:related-work}.

\begin{algorithm}[!h]
	\caption{A farthest first search (FFS) algorithm for exemplar selection}
	\label{alg:ffs}
	\begin{algorithmic}[1]
		\REQUIRE Data $\cX = \{\x_1, \dots, \x_N\} \subseteq \Re ^D$, parameter $\lambda > 1$ and number of desired exemplars $k \ll N$.
		\STATE Select $j\in\{1,\dots,N\}$ randomly and set 
		$\cX_0^{(1)} \gets \{\x_j\}$.
		\FOR {$i=1, \cdots, k-1$}
		\STATE \label{step:update}$\cX_0^{(i+1)} = \cX_0^{(i)} ~ \cup ~ \argmax_{\x_j \in \cX} f_\lambda(\x_j, \cX_0^{(i)})$
		\ENDFOR
		\ENSURE $\cX_0^{(k)}$
	\end{algorithmic}
\end{algorithm}


\begin{algorithm}[!h]
	\caption{An efficient implementation of FFS}
	\label{alg:ffs-efficient}
	\begin{algorithmic}[1]
		\REQUIRE Data $\cX = \{\x_1, \dots, \x_N\} \subseteq \Re ^D$, parameters $\lambda > 1$ and number of desired exemplars $k \ll N$.
		\STATE Select $j\in\{1,\dots,N\}$ randomly and set $\cX_0^{(1)} \gets\{\x_j\}$.
		\STATE \label{step:init_b}Compute $b_j = f_\lambda(\x_{j}, \cX_0^{(1)})$ for $j = 1, \cdots, N$. 
		\FOR {\label{step:outer}$i=1, \cdots, k-1$}
		\STATE \label{step:order}Let $o_1, \cdots, o_N$ be a permutation of  $1, \cdots, N$ such that 
		$$
		b_{o_p} \ge b_{o_q} \ \ \text{when} \ \ p < q.
		$$
		\STATE Initialize \emph{max\_cost}  $=0$.
		\FOR {$j=1, \cdots, N$}
		\STATE \label{step:update_b}Set $b_{o_j} = f_\lambda(\x_{o_j}, \cX_0^{(i)})$.
		\IF {$b_{o_j} > $ \emph{max\_cost}}
		\STATE \label{step:max}Set \emph{max\_cost} $= b_{o_j}$, \emph{new\_index} $=o_j$.
		\ENDIF
		\IF {\label{step:break_condition}$j = N$ or \emph{max\_cost} $\ge b_{o_{j+1}}$}
		\STATE \label{step:break} \textbf{break}
		\ENDIF
		\ENDFOR
		\STATE $\cX_0^{(i+1)} = \cX_0^{(i)} ~ \cup ~ \{\x_\text{\emph{new\_index}}\}$.
		\ENDFOR
		\ENSURE $\cX_0^{(k)}$
	\end{algorithmic}
\end{algorithm}

\myparagraph{Efficient implementation} 
Observe that each iteration of Algorithm~\ref{alg:ffs} requires evaluating $f_\lambda(\x_j, \cX_0^{(i)})$ for every $\x_j \in \cX$. 
Therefore, the complexity of Algorithm~\ref{alg:ffs} is linear in the number of data points $N$ assuming $k$ is fixed and small. 
However, computing $f_\lambda(\x_j, \cX_0^{(i)})$ itself is not easy as it requires solving a sparse optimization problem. 
Next, we introduce an efficient implementation of Algorithm~\ref{alg:ffs} that accelerates the procedure by eliminating the need to compute $f_\lambda(\x_j, \cX_0^{(i)})$ for some $\x_j$ in each iteration. 

The idea underpinning the computational savings in Algorithm~\ref{alg:ffs-efficient} is the monotonicity of $f_\lambda(\x_j, \cdot)$ (see Lemma~\ref{thm:f-monotone}). 
That is, for any $\emptyset \subseteq \cX_0' \subseteq \cX_0'' \subseteq \cX$ we have $f_\lambda(\x_j, \cX_0') \ge f_\lambda(\x_j, \cX_0'')$.
Since in the FFS algorithm the set $\cX_0^{(i)}$ is progressively increased, this implies that $f_\lambda(\x_j, \cX_0^{(i)})$ is non-increasing in $i$.
In step~\ref{step:init_b} we initialize $b_j=f_\lambda(\x_j, \cX_0^{(1)})$ for each $j\in \{1, \cdots, N\}$, which is an upper bound for $f_\lambda(\x_j, \cX_0^{(i)})$ for $i \ge 1$. In each iteration $i$, the goal is to find a data point that maximizes $f_\lambda(\x_j, \cX_0^{(i)})$. 
To do this, we first find an ordering $o_1, \cdots, o_N$ of $1, \cdots, N$ such that $b_{o_1} \ge  \cdots \ge b_{o_N}$ (step~\ref{step:order}). 
We then compute $f_\lambda(\cdot, \cX_0^{(i)})$ sequentially for points in $\x_{o_1}, \cdots, \x_{o_N}$ (step~\ref{step:update_b}) while tracking the highest value of $f_\lambda(\cdot, \cX_0^{(i)})$ by the variable \emph{max\_cost} (step~\ref{step:max}).
Once the condition that \emph{max\_cost} $\ge b_{o_{j+1}}$ is met (step~\ref{step:break_condition}), we can assert that for any $j' > j$ the point $\x_{o_{j'}}$ is not a maximizer. This can be seen from $f_\lambda(\x_{o_{j'}}, \cX_0^{(i)}) \le b_{o_{j'}} \le b_{o_{j+1}} \le$ \emph{max\_cost}, where the first inequality follows from the monotonicity of $f_\lambda(\x_{o_{j'}}, \cX_0^{(i)})$ as a function of $i$.
Thus, we can break the loop (step~\ref{step:break}) and avoid computing $f_\lambda(\x_{o_{j}}, \cX_0^{(i)})$ for the remaining values of $j$ in this iteration.
When Algorithm~\ref{alg:ffs-efficient} terminates, it produces the same output as  Algorithm~\ref{alg:ffs} but with a reduced total number of evaluations for $f_\lambda(\cdot, \cdot)$. 

\begin{figure}[!htb]
	\centering
	\includegraphics[scale=0.7]{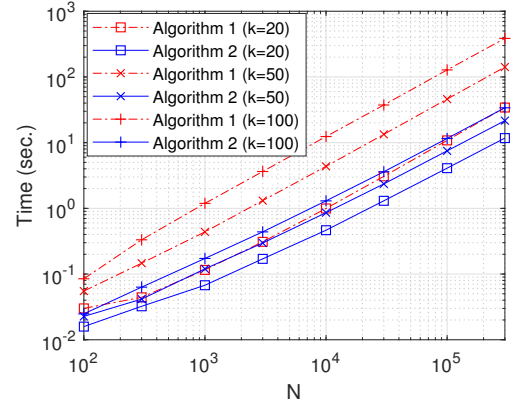}
	\caption{\label{fig:efficient} Running time for Algorithm~\ref{alg:ffs} and Algorithm~\ref{alg:ffs-efficient} on a synthetically generated dataset where $N$ data points are sampled uniformly at random from the unit sphere of $\RR^{10}$ averaged over $10$ trials. $N$ is varied along the x-axis and takes values between $100$ and $300,\!000$. }
\end{figure}

{
Figure~\ref{fig:efficient} reports the computational time of Algorithm~\ref{alg:ffs} and Algorithm~\ref{alg:ffs-efficient} with synthetically generated data where $N$ data points are sampled uniformly at random from the unit sphere of $\RR^{10}$. 
It shows that the efficient implementation in Algorithm~\ref{alg:ffs-efficient} is around $2$ to $10$ times faster than the naive implementation in Algorithm~\ref{alg:ffs}.
Comparing the results across different values of $k$, we find that the benefit of Algorithm~\ref{alg:ffs-efficient} is more prominent for larger values of $k$.
}


\section{Theoretical Analysis}
\label{sec:ESC_theory}

In this section, we study the theoretical properties of the self-representation based exemplar selection method. 
In Section~\ref{sec:theory-geometry} and~\ref{sec:theory-union-of-subspaces} we present a geometric interpretation of the exemplar selection model from Section~\ref{sec:ESC_model} and the FFS algorithm from Section~\ref{sec:FFS}, and study their properties when data is drawn from a union of subspaces.
To simplify the analysis, we assume that the self-representation $\x_j=\sum_{i \ne j} c_{i} \x_i$ is strictly enforced by extending \eqref{eq:representation-f} to $\lambda = \infty$, i.e., we let 
\begin{equation}
f_\infty(\x_j, \cX_0) = \min_{\c \in \Re^N} \|\c\|_1 \st \x_j = \sum_{i: \x_i \in \cX_0} c_{i} \x_i.
\label{eq:representation-f_infinity}
\end{equation}
We define $f_\infty(\x_j, \cX_0) = \infty$ if problem~\eqref{eq:representation-f_infinity} is infeasible. 
{The effect of using a finite $\lambda$ is discussed in Section~\ref{sec:theory-lambda}.}

\subsection{Geometric interpretation}\label{sec:theory-geometry}
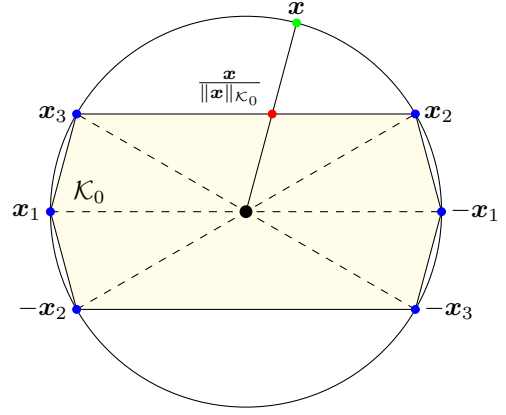
\begin{figure}
	\centering
	\def\angleF{0}
	\def\angleS{30}
	\def\angleT{180 - \angleS}
	\def\angle{75}
	
	\begin{tikzpicture}[scale = 2.6]
	\coordinate (0) at (0,0);
	\draw[black, fill = none] (0) circle [radius = 1];
	\draw[fill=yellow!10] (\angleF:1 cm) -- (\angleS:1 cm) -- (\angleT:1 cm) -- (\angleF + 180:1 cm) -- (\angleS + 180:1 cm) -- (\angleT + 180:1 cm) -- cycle;
	\node at (-0.8, 0.1) {$\cK_0$};
	\draw [black, fill=black] (0) circle [radius=0.03];
	\draw [dashed] (\angleF+180:1 cm) -- (\angleF:1 cm);
	\draw [dashed] (\angleS+180:1 cm) -- (\angleS:1 cm);
	\draw [dashed] (\angleT+180:1 cm) -- (\angleT:1 cm);
	\draw [blue, fill=blue] (\angleF:1 cm) circle [radius=0.02];
	\draw [blue, fill=blue] (180+\angleF:1 cm) circle [radius=0.02];
	\draw [blue, fill=blue] (\angleS:1 cm) circle [radius=0.02];
	\draw [blue, fill=blue] (180+\angleS:1 cm) circle [radius=0.02];
	\draw [blue, fill=blue] (\angleT:1 cm) circle [radius=0.02];
	\draw [blue, fill=blue] (180+\angleT:1 cm) circle [radius=0.02];
	\node[left] at (\angleF + 180:1 cm) {$\x_1$};
	\node[right] at (\angleF:1 cm) {$-\x_1$};
	\node[left] at (\angleS + 180:1 cm) {$-\x_2$};
	\node[right] at (\angleS:1 cm) {$\x_2$};
	\node[right] at (\angleT + 180:1 cm) {$-\x_3$};
	\node[left] at (\angleT:1 cm) {$\x_3$};
	\draw [solid] (0, 0) -- (\angle:1 cm);
	\draw [green, fill=green] (\angle:1 cm) circle [radius=0.02];
	\node[above] at (\angle:1 cm) {$\x$};
	\draw [red, fill=red] (\angle:0.5176 cm) circle [radius=0.02];
	\node[above left] at (\angle:0.5176 cm) {$\frac{\x}{\|\x\|_{\cK_0}}$};
	\end{tikzpicture}
	\caption[A geometric illustration of the solution to \eqref{eq:exemplar-objective}]{\label{fig:geometry}A geometric illustration of the solution to \eqref{eq:exemplar-objective} with $\cX_0=\{\x_1, \x_2, \x_3\}$. The shaded area is the convex hull $\cK_0$ defined in~\eqref{def:K0}.}
\end{figure}

We first provide a geometric interpretation of the exemplars selected by \eqref{eq:exemplar-objective}. 
Given any $\cX_0$, we denote the convex hull of the symmetrized data points in $\cX_0$ by $\cK_0$, i.e., 
\begin{equation} \label{def:K0}
\cK_0 := \text{conv}(\pm \cX_0)
\end{equation}
(see an example in Figure~\ref{fig:geometry}).
The Minkowski functional~\cite{Vershynin:09} associated with a set $\cK_0$ is given by the following.

\begin{definition}[Minkowski functional]
	\label{def:Minkowski}
	The Minkowski functional associated with a set $\cK_0 \subseteq \Re^D$ is a map denoted by $\|\cdot\|_{\cK_0}: \Re^D \to \Re \cup \{+\infty\}$ and defined by
	\begin{equation}
	\|\x\|_{\cK_0}:= \inf\{t>0: \x /t \in \cK_0\}.
	\label{eq:Minkowski}
	\end{equation}
	We define $\|\x\|_{\cK_0}:=\infty$ if $\{t>0: \x /t \in \cK_0\}$ is empty.
\end{definition}


The Minkowski functional is a norm on $\spann(\cK_0)$, 
and its unit ball is $\cK_0$.
Thus, for any nonzero $\x \in \spann(\cK_0)$, the point $\x / \|\x\|_{\cK_0}$ is the projection onto the boundary of $\cK_0$. The green and red dots in Figure~\ref{fig:geometry} are examples of $\x$ and $\x / \|\x\|_{\cK_0}$, respectively. 
It follows that if $\|\x\|_2 = 1$, then $1/ \|\x\|_{\cK_0}$ is the length of the ray $\{t\x: t \ge 0\}$ inside $\cK_0$.

Using Definition~\ref{def:Minkowski}, it has been shown by \cite[Section 2]{Donoho:TechRep05}\cite[Section 4.1]{Soltanolkotabi:AS12} that
\begin{equation}\label{eq:basis-pursuit-Minkowski}
\|\x\|_{\cK_0} = f_\infty(\x, \cX_0) \  \ \text{for all} \ \ \x \in \Re^D.
\end{equation}
A combination of \eqref{eq:basis-pursuit-Minkowski} and the interpretation of $1/ \|\x\|_{\cK_0}$ above provides a geometric interpretation of $f_\infty(\x, \cX_0)$.
That is, $f_\infty(\x, \cX_0)$ is large if the length of the ray $\{t\x: t \ge 0\}$ inside $\cK_0$ is small.
In particular, it holds that $f_\infty(\x, \cX_0)$ is infinity if $\x$ is not in the span of $\cX_0$.

In view of \eqref{eq:basis-pursuit-Minkowski}, the exemplar selection model \eqref{eq:exemplar-objective} may be written equivalently as 
\begin{equation}
\cX_0^* = \argmax_{|\cX_0| \le k} \inf_{\x_j \in \cX}1 / \|\x_j\|_{\cK_0}. 
\end{equation}
Therefore, the solution to \eqref{eq:exemplar-objective} is the subset $\cX_0$ of $\cX$ that maximizes where the ray $\{t\x_j: t \ge 0\}$ intersects $\cK_0$ taken over all  data $\x_j \in \cX$ (i.e., maximizes the minimum of such intersections over all $\x_j\in\cX$).

Also, from \eqref{eq:basis-pursuit-Minkowski} we  see that each iteration of Algorithm~\ref{alg:ffs} selects the $\x_j$ that minimizes $1 / \|\x_j\|_{\cK_0}$.
Therefore, each iteration of FFS adds the point $\x_j \in \cX$ whose associated ray $\{t\x_j: t>0\}$ has the shortest  intersection with $\cK_0$.

{
Finally, we remark that our exemplar selection objective is related to the sphere covering problem. This is discussed in detail in the Appendix. 
}

\subsection{Exemplars from a union of subspaces}\label{sec:theory-union-of-subspaces}

We now study the properties of our exemplar selection method when applied to data from a union of subspaces. 
Let $\cX$ be drawn from a collection of subspaces $\{\cS_\ell\}_{\ell=1}^n$ of dimensions $\{d_\ell\}_{\ell=1}^n$ with 
each subspace $\cS_\ell$ containing at least $d_\ell$ samples that span $\cS_\ell$. 
We assume that the subspaces are independent, which is commonly used in the analysis of subspace clustering methods \cite{Vidal:IJCV08,Elhamifar:CVPR09,Lu:ECCV12,Liu:ICML10,You:CVPR16-SSCOMP}.
\begin{assumption}
	The subspaces $\{\cS_\ell\}_{\ell=1}^n$ are independent, i.e.,  $\sum_{\ell=1}^n d_\ell$ is equal to the dimension of $\sum_{\ell=1}^n \cS_\ell$.
	\label{asmp:independent}
\end{assumption}

We now aim to show that the solution to \eqref{eq:exemplar-objective} contains at least $d_\ell$ independent vectors from each subspace $\cS_\ell$ and, moreover,
the solution to \eqref{eq:esc} with $\cX_0$ being any solution to \eqref{eq:exemplar-objective} is subspace-preserving for all $j\in\{1,\dots,N\}$. 
Formally, the subspace-preserving property is defined as follows.
\begin{definition}[Subspace-preserving property]
    A vector $\c\in \Re^N$ associated with $\x_j \in\cX$ is called subspace-preserving if $c_i \ne 0$ implies that $\x_i$ and $\x_j$ are from the same subspace.
\end{definition}



We first need the following lemma.
\begin{lemma}\label{thm:independent-lemma}
	Suppose that $\x_j \in \cS_\ell$. Under Assumption~\ref{asmp:independent}, 
	if the  optimization problem in \eqref{eq:representation-f_infinity} is feasible, then
	any optimal solution $\c^*$ to it satisfies
	$\x_j = \sum_{i: \x_i \in \cX_0 \cap \cS_\ell} c_i^* \x_i$, and $c_i^* = 0$ for all $i$ satisfying $\x_i\notin \cX_0\cap \cS_\ell$,
	i.e., $\x_j$ is expressed as a linear combination of points in $\cX_0$ that are from its own subspace.
\end{lemma}
\begin{proof}
	An optimal solution $\c^*$ to~\eqref{eq:representation-f_infinity} must be feasible, i.e.,
	\begin{equation*}
	\x_j = \sum_{i: \x_i \in \cX_0}c_i^* \x_i = \sum_{i: \x_i \in {\cX_0}\cap\cS_\ell}c_i^*\x_i + \sum_{m \ne \ell} 
	\Big( \sum_{i: \x_i \in {\cX_0}\cap\cS_m}c_i^*\x_i\Big),
	\end{equation*} 
	which after rearrangement gives
	\begin{equation}
	\x_j - \sum_{i: \x_i \in {\cX_0}\cap\cS_\ell}c_i^*\x_i = \sum_{m \ne \ell}
	\Big(\sum_{i: \x_i \in {\cX_0}\cap\cS_m}c_i^*\x_i\Big).
	\end{equation}
	Since the left-hand side is a vector in $\cS_\ell$ and the right-hand side is a vector in $\sum_{m\ne \ell}\cS_m$, it follows from Assumption~\ref{asmp:independent} and~\cite[Theorem~6]{hoffman1971linear} that
	$\x_j = \sum_{i: \x_i \in {\cX_0}\cap\cS_\ell}c_i^*\x_i$, as claimed.
	
	Next, let us define the vector $\hat{\c}$ such that $\hat{c}_i = c^*_i$ for all $i: \x_i \in {\cX_0} \cap \cS_\ell$ and
	$\hat{c}_i = 0$ for all $i$ such that $\x_i\notin \cX_0\cap\cS_\ell$.
	Using $\x_j = \sum_{i: \x_i \in {\cX_0}\cap\cS_\ell}c_i^*\x_i$ from above and the definition of $\hat{c}$, we see that $\hat{c}$ is  
	feasible for \eqref{eq:representation-f_infinity}. 
	Moreover, it satisfies
	\begin{equation}\label{eq:prf-feasible-solution}
	\|\hat{\c}\|_1 
	= \sum_{i:\x_i \in {\cX_0}\cap \cS_\ell} |c^*_i| 
	\leq \|\c^*\|_1.
	\end{equation}
	Since $\c^*$ is optimal for  \eqref{eq:representation-f_infinity}, it follows from  \eqref{eq:prf-feasible-solution} that $\|\hat{\c}\|_1 = \|\c^*\|_1$. Combining this fact with \eqref{eq:prf-feasible-solution} shows that $c^*_i = 0$ for all $i$ such that $\x_i\notin\cX_0\cap\cS_\ell$, which completes the proof.
\end{proof}

We may use this lemma to prove the following result.

\begin{theorem}
	Under Assumption~\ref{asmp:independent}, for all $k \ge \sum_{\ell=1}^n d_\ell$, any solution $\cX^*_0$ to the optimization problem \eqref{eq:exemplar-objective} contains at least $d_\ell$ linearly independent points from each subspace $\cS_\ell$. 
	Moreover, with $\cX_0 = \cX^*_0$, the optimization problem in~\eqref{eq:representation-f_infinity} is feasible for all $\x_j\in\cX$ with all optimal solutions
	being subspace-preserving.
	\label{thm:subset-from-subspace}
\end{theorem}
\begin{proof}
    Let $\cX_0^*$ be any optimal solution to \eqref{eq:exemplar-objective} for any fixed $k \ge \sum_{\ell=1}^n d_\ell$, and $\cX_0 \subseteq \cX$ be any subset with $|\cX_0|=k$ that contains $d_\ell$ linearly independent points from $\cS_\ell$ for each $\ell \in\{1, \cdots, n\}$, which we know exists.
	It follows that $f_\infty(\x_j, \cX_0) < \infty$ for all $\x_j \in \cX$ so that $F_\infty(\cX_0) < \infty$. This and optimality of $\cX_0^*$ imply $F_\infty(\cX_0^*) \le F_\infty(\cX_0) < \infty$. This fact and the definition of $F_\infty(\cX_0^*)$ means that $f_\infty(\x_j,\cX_0^*) < \infty$ for all $j$, i.e., $\x_j\in\spann(\cX_0^*)$ for all $j$.  Combining this with Assumption~\ref{asmp:independent} implies that $\cX_0^*$ contains at least $d_\ell$ linearly independent points from each subspace $\cS_\ell$, which also  means that the problem in~\eqref{eq:representation-f_infinity} is feasible for all $\x_j\in\cX$. Combining this with 
Lemma~\ref{thm:independent-lemma} shows that all solutions to the optimization problem in~\eqref{eq:representation-f_infinity} are subspace preserving. 
	%
\end{proof}

When $k = \sum_{\ell=1}^n d_\ell$, 
Theorem~\ref{thm:subset-from-subspace} shows that $d_\ell$ points are selected from subspace $\cS_\ell$ regardless of the number of points in $\cS_\ell$. 
Therefore, when the data is class imbalanced, \eqref{eq:exemplar-objective} selects a subset that is more balanced provided the dimensions of the subspaces do not differ dramatically.  

Theorem~\ref{thm:subset-from-subspace} also shows that only  $\sum_{\ell=1}^n d_\ell$ points are needed to correctly represent all data points in $\cX$. In other words, the required number of exemplars for representing the dataset does not scale with the size of the dataset $\cX$.

Although the FFS algorithm in Section~\ref{sec:FFS} is a computationally efficient greedy algorithm that does not necessarily solve \eqref{eq:exemplar-objective}, the following result shows that it does output a subset of exemplars from the data with desirable properties. 

\begin{theorem}
	The conclusions of Theorem~\ref{thm:subset-from-subspace} hold when $\cX_0^*$ is replaced by 
	$\cX_0^{(k)}$ for any $k \geq \sum_{\ell=1}^n d_\ell$, where $\cX_0^{(k)}$ is the set of exemplars returned by Algorithm~\ref{alg:ffs}
	\label{thm:ffs} (equivalently,  Algorithm~\ref{alg:ffs-efficient}).
\end{theorem}

\begin{proof}
Note that since $\lambda = \infty$, it follows from the definition in~\eqref{eq:representation-f_infinity} that $f_\infty(x_j, \cX_0^{(i)}) = \infty$ if and only if $x_j \notin \spann(\cX_0^{(i)})$. It follows from this fact and the construction of Algorithm~\ref{alg:ffs} that each iteration $i$ of Algorithm~\ref{alg:ffs} adds a data point from $\cX$ that is linearly independent from those in $\cX_0^{(i)}$ provided such a linearly independent vector exists.  However, we know from Assumption~\ref{alg:ffs} that there exists $\bar{k} := \sum_{\ell=1}^n d_\ell \leq k$ linearly independent vectors in $\cX$.  Putting this all together means that $\cX_0^{\bar{k}}$ will contain exactly $d_\ell$ linearly independent points from each subspace $\cS_\ell$, which also means that the optimization problem in~\eqref{eq:representation-f_infinity}  is feasible for all $\x_j\in\cX$. 
 Combining this with  Lemma~\ref{thm:independent-lemma} completes the proof. 
\end{proof}


{
\subsection{Effect of the regularization parameter $\lambda$}\label{sec:theory-lambda}
The analysis in Section~\ref{sec:theory-geometry} and~\ref{sec:theory-union-of-subspaces} concentrates on the case where the regularization parameter $\lambda$ is set to $\infty$. 
In real applications where the data $\cX$ contains noise and thus deviates from the union-of-subspace model, the self-representation constraint $\x_j=\sum_{i \ne j} c_{i} \x_i$ may not be strictly satisfied. 
In such cases, using a finite $\lambda$ makes sense.

On the other hand, the value of $\lambda$ should also not be too small. 
The following theorem states that if $\lambda$ is below a certain threshold then the value $f_\lambda(\x_j, \cX_0)$ is the same for all $\x_j \in \cX$, and therefore no longer provides a measure of how well the data point $\x_j$ is represented by $\cX_0$. 
Consequently, Algorithm~\ref{alg:ffs} and Algorithm~\ref{alg:ffs-efficient} will fail to produce a useful representative subset of exemplars. 

\begin{theorem}
	Given any $\cX_0 \subseteq \cX$, we have $f_\lambda(\x_j, \cX_0) = \frac{\lambda}{2}$ for all $\x_j \in \cX \setminus \cX_0$ if
	\begin{equation}\label{eq:lambda_bound}
		\lambda < \frac{1}{\max\limits_{\x' \in \cX}\max\limits_{\x'' \in \cX, \x'' \ne \x'}|\langle \x', \x'' \rangle|}.
	\end{equation}
\end{theorem}
\begin{proof}
	The optimality condition for the optimization problem in \eqref{eq:representation-f} is given by
	\begin{equation}
	    \lambda \x_i^\transpose (\x_j - \sum_{i:\x_i \in \cX_0}c_i \x_i) \in \partial |c_i|, ~~\forall i:\x_i \in \cX_0,
	\end{equation}
	which is satisfied by $\c = \0$ when \eqref{eq:lambda_bound} is satisfied.
	Therefore, $\c = \0$ is an optimal solution to \eqref{eq:representation-f}. 
	Plugging this solution into the objective function of \eqref{eq:representation-f} gives $f_\lambda(\x_j, \cX_0) = \frac{\lambda}{2}$.
\end{proof}

}

\section{The Application of Exemplar Selection to Subspace Clustering and Classification}
\label{sec:assignment}

In this section, we present procedures for using the exemplars returned by Algorithm~\ref{alg:ffs-efficient} to generate class assignments for data drawn from a union of subspaces.
In Section~\ref{sec:esc} we consider the problem of subspace clustering where the class labels of all data are unknown. In Section~\ref{sec:es4c} we consider a setting where the class labels for the exemplars are obtained, and then used to  classify the remaining data points.

\subsection{Exemplar based subspace clustering} 
\label{sec:esc}

Once a set of exemplars $\cX_0$ has been generated, we can compute a representation vector $\c_j$ for each $\x_j \in \cX$ as the solution to the optimization problem~\eqref{eq:esc}.
As shown in Theorem~\ref{thm:ffs}, the vector $\c_j$ is expected to be subspace-preserving, i.e., $c_{ij}$ is nonzero only if $\x_i$ and $\x_j$ are from the same subspace.
Motivated by this observation, we use a nearest neighbor approach to compute the segmentation of $\cX$ (see Algorithm~\ref{alg:esc}). 
First, the coefficient vectors $\{\c_j\}$ are normalized, i.e., we set $\tilde{\c}_j = \c_j / \|\c_j\|_2$. 
Then, for each $\tilde{\c}_j$ we find $t$-nearest neighbors with  the largest positive inner product with $\tilde{\c}_j$. 
Note that if all vectors in $\{\c_j\}_{j=1}^N$ are subspace-preserving, then for any two points $\{\x_i, \x_j\} \subseteq \cX$ we have $\langle \tilde{\c}_i, \tilde{\c}_j\rangle >  0$ only if $\x_i$ and $\x_j$ are from the same subspace.
Therefore, the $t$-nearest neighbors of $\tilde{\c}_j$ from this step all come from the same subspace as $\x_j$.
Finally, we compute an affinity matrix from the $t$-nearest neighbors and apply spectral clustering to get the segmentation\footnote{While there could be many other procedures for generating class assignments from the coefficient vectors $\{\c_j\}$ such as those in \cite{Elhamifar:TPAMI13,Adler:TNNLS15,Traganitis:TSP17}, we find in our numerical experiments that the procedure described in Algorithm~\ref{alg:esc} usually works at least as well.}.

\begin{algorithm}[h]
	\caption{Exemplar based subspace clustering (ESC)}
	\label{alg:esc}
	\begin{algorithmic}[1]
		\REQUIRE Data $\cX = \{\x_1, \dots, \x_N\} \subseteq \Re ^D$, parameter $\lambda > 1$, number of exemplars $k$ and number of neighbors $t$.
		\STATE \label{step:esc-ffs} Compute  exemplars $\cX_0 = \cX_0^{(k)}$ using Algorithm~\ref{alg:ffs-efficient}. Then compute $\{\c_j\}_{j=1}^N$ with $\c_j$ a solution of  \eqref{eq:esc}.
		\STATE \label{step:knn}Define $\tilde{\c}_j = \c_j / \|\c_j\|_2$ for all $j$. For all $i$ and $j$, set $\W_{ij}=1$ if $\tilde{\c}_j$ is a $t$-nearest neighbor of $\tilde{\c}_i$ and $\langle \tilde{\c}_j, \tilde{\c}_i \rangle > 0$, and $\W_{ij}=0$ otherwise. 
		\STATE \label{step:affinity}Set $\A=\W+\W^\top$ and apply spectral clustering to $\A$.
		\ENSURE Segmentation of $\cX$.
	\end{algorithmic}
\end{algorithm}

\begin{theorem}
	Take any $k\ge\sum_{\ell=1}^n d_\ell$ and any $t > 0$. Let $\lambda = \infty$. Under Assumption~\ref{asmp:independent}, the affinity matrix $\A$ in step~\ref{step:affinity} of Algorithm~\ref{alg:esc} has no wrong connections, i.e., the $(i,j)$-th entry of $\A$ is nonzero only if $\x_i$ and $\x_j$ are from the same subspace.
	\label{thm:esc-correct-clustering}
\end{theorem}
\begin{proof}
    From Theorem~\ref{thm:ffs} we know that the vectors in $\{\c_j\}_{j=1}^N$ computed in step~\ref{step:esc-ffs} of Algorithm~\ref{alg:esc} are subspace-preserving. Therefore, $\langle \tilde{\c}_j, \tilde{\c}_i \rangle > 0$ only if $\x_i$ and $\x_j$ are from the same subspace. 
    Then, according to the steps for computing $\W$ and $\A$ in Algorithm~\ref{alg:esc} we know that the $(i,j)$-th entry of $\A$ is nonzero only if $\x_i$ and $\x_j$ are from the same subspace.
\end{proof}
By Theorem~\ref{thm:esc-correct-clustering}, each nonzero entry in the affinity matrix $\A$ corresponds to pairs of points that are in the same subspace. Although this conclusion holds for
all $t > 0$, if the value for $t$ is chosen to be too small, then
data points from the same subspace will not form a single connected component in the associated graph, leading to the issue of over-segmentation.
In our experiments, we find that $t$ needs to be at least three to produce good clustering performance.

\subsection{Exemplar selection for subspace classification}
\label{sec:es4c}

Given a large-scale unlabeled dataset, it is expensive to manually annotate all data.
One remedy is to select a small subset of data for manual labeling, and then infer the labels for the remaining data by training a model on the selected subset.
In the following, we assume that the exemplars selected by Algorithm~\ref{alg:ffs-efficient} have been labeled, and present the sparse representation based classification \cite{Wright:PAMI09} technique to classify the rest of the data points (see Algorithm~\ref{alg:es4c}). 
For each data point $\x_j$, we compute the reconstruction residual with respect to each class $\ell$ as $\r_j^{(\ell)}:=\x_j - \sum_{i: \x_i \in \cX_0^{(\ell)}}c_{ij}\x_i$, where as above $\c_j$ is computed as the solution to~\eqref{eq:esc} for each $j$,  and
$\cX_0^{(\ell)}$ denotes the subset of the exemplars that are from class $\ell$.
Note that if $\c_j$ is subspace-preserving, then $\x_j$ can be  represented by exemplars from its own class with zero  reconstruction residual. In practice, we expect
that $\|\r_j^{(\ell)}\|_2 \ll \|\x_j\|_2$ for the class $\ell$ that $\x_j$ belongs to, and that $\|\r_j^{(\ell)}\|_2 = \|\x_j\|_2$ for all other classes. 
Motivated by this observation, we choose to assign $\x_j$ to the class that gives the minimum reconstruction residual.

\begin{algorithm}[!h]
	\caption{Exemplar selection for subspace classification}
	\label{alg:es4c}
	\begin{algorithmic}[1]
		\REQUIRE Data $\cX = \{\x_1, \dots, \x_N\} \subseteq \Re ^D$, parameter $\lambda > 1$, and number of exemplars $k$.
		\STATE \label{step:es4c-ffs} Compute exemplars $\cX_0 = \cX_0^{(k)}$ from Algorithm~\ref{alg:ffs-efficient}.  Then compute $\{\c_j\}_{j=1}^N$ with $\c_j$ a solution to~\eqref{eq:esc}.
		\STATE Request the class label of points in $\cX_0$. Define $\cC_0^{(\ell)} \subseteq \cX_0$ as the subset of exemplars from class $\ell$.
		\STATE \label{step:es4c-assign}Assign each $\x_j\in \cX\setminus\cX_0$ to the class that solves the problem $\argmin_\ell \|\x_j - \sum_{i: \x_i \in \cC_0^{(\ell)}}c_{ij}\x_i\|_2$.
		\ENSURE Segmentation of $\cX$.
	\end{algorithmic}
\end{algorithm}

The following theorem shows that the output of Algorithm~\ref{alg:es4c} is a correct segmentation of the data $\cX$.

\begin{theorem}
	Take any $k\ge\sum_{\ell=1}^n d_\ell$, and let $\lambda = \infty$. Under Assumption~\ref{asmp:independent}, the output of Algorithm~\ref{alg:es4c} is such that each point in $\cX$ has the correct class label, i.e., the segmentation is correct.
\end{theorem}
\begin{proof}
    Note that in Algorithm~\ref{alg:es4c} we have $\cX_0 = \cX_0^{(k)}$, where $\cX_0^{(k)}$ is the set of exemplars  returned by Algorithm~\ref{alg:ffs-efficient}.  Now, consider $\x_j \in \cX \backslash \cX_0$, and assume without loss of generality that $\x_j \in \cS_\ell$.
    From Theorem~\ref{thm:ffs}, the vector $\c_j$ computed in step~\ref{step:es4c-ffs} of Algorithm~\ref{alg:es4c} is subspace-preserving so that
    \begin{equation}\label{wrong-class}
        \|\x_j - \sum_{i: \x_i \in \cC_0^{(p)}}c_{i}\x_i\|_2 
        = \|\x_j\|_2 
        = 1
        \ \ \text{for all $p\neq \ell$}
    \end{equation}
    and
    \begin{equation}\label{right-class}
        \x_j - \sum_{i: \x_i \in \cC_0^{(\ell)}}c_{i}\x_i = \0.
    \end{equation}
    From~\eqref{wrong-class}, \eqref{right-class}, and  step~\ref{step:es4c-assign} of Algorithm~\ref{alg:es4c}, it follows that the point $\x_j$ is assigned to its correct class, namely $\ell$.
\end{proof}

\section{Experiments}
\label{sec:esc-experiments}

In this section, we demonstrate the performance of our exemplar selection method for subspace clustering and subspace classification tasks. 
The sparse optimization problem~\eqref{eq:esc} that must be solved to perform 
step~\ref{step:update_b} of Algorithm~\ref{alg:ffs-efficient},
step~\ref{step:esc-ffs} of Algorithm~\ref{alg:esc}, and step~\ref{step:es4c-ffs} of Algorithm~\ref{alg:es4c}
is solved by the LASSO version of the LARS algorithm \cite{Efron:AS04} implemented in the SPAMS package~\cite{Mairal:JMLR2010}. 
The nearest neighbors in step~\ref{step:knn} of Algorithm~\ref{alg:esc} are computed by the $k$-d tree algorithm implemented in the VLFeat toolbox \cite{vedaldi:vlfeat}. 

\myparagraph{Databases} We use three publicly available databases. 
The Extended MNIST (EMNIST) dataset \cite{Cohen:arXiv17} is an extension of the MNIST dataset that contains gray-scale handwritten digits and letters. 
We take all $190,\!998$ images corresponding to $26$ lower case letters, and use them as the data for a $26$-class clustering problem. 
The size of each image in this dataset is $28$ by $28$. 
Following \cite{You:CVPR16-SSCOMP}, each image is represented by a feature vector computed from a scattering convolutional network \cite{Bruna:PAMI13}, which is translational invariant and deformation stable (i.e. it linearizes small deformations). Therefore, these features from EMNIST approximately follow a union of subspaces model.

The German Traffic Sign Recognition Benchmark (GTSRB) database \cite{Stallkamp:NN12}  contains $43$ categories of street sign data with over $50,\!000$ images in total. We remove categories associated with speed limit and triangle-shaped signs (except the yield sign) as they are difficult to distinguish from each other, which results in a final data set of $12,\!390$ images in $14$ categories. Each image is represented by a $1,\!568$-dimensional HOG feature \cite{Dalal:CVPR05} provided with the database.
The main intra-class variation in GTSRB is the illumination conditions, therefore the data can be well-approximated by a union of subspaces \cite{Basri:PAMI03}.

For both EMNIST and GTSRB, feature vectors are mean subtracted and projected to dimension $500$ by PCA and normalized to have unit $\ell_2$ norm.
Both the EMNIST and GTSRB databases are imbalanced. In EMNIST, for example, the number of images for each letter ranges from $2,\!213$ (letter ``j'') to $28,\!723$ (letter ``e''), and the number of samples for each letter is approximately equal to their frequencies in the English language. In Figure~\ref{fig:frequency} we show the number of instances for each class in both of these databases.

In order to compare with other methods that are not able to handle large scale datasets, we create a small scale imbalanced dataset from the Extended Yale B face database \cite{Kriegman:PAMI01}. 
The Extended Yale B face database contains images of $38$ faces and each of them is taken under $64$ different illumination conditions. 
We randomly select $10$ classes and sample a subset from each class. The number of images
we sample for those $10$ classes is $16$ for the first three classes, $32$ for the next three classes and $64$ for the remaining four classes.
The images are preprocessed by standardization (i.e., the images are subtracted by the mean image and divided by the standard deviation) and subsequently normalized to have unit $\ell_2$ norm for all methods except for the separable NMF methods as they require nonnegative input.

\subsection{Exemplar based subspace clustering}
\label{sec:esc-experiments-subspace-clustering}

We demonstrate the performance of our exemplar subspace clustering Algorithm~\ref{alg:esc} (henceforth referred to as ESC-FFS) for subspace clustering on class-imbalanced databases. 
We set $\lambda$ to $150$, $15$ and $100$ for EMNIST, GTSRB and Extended Yale B, respectively, and set $t$ to $3$ for all three databases.

\myparagraph{Baselines} 
We compare our approach with SSC \cite{Elhamifar:TPAMI13} to show the effectiveness of exemplar selection in addressing imbalanced data. 
For solving the sparse recovery problem in SSC, we use the algorithm in \cite{You:CVPR16-EnSC} which is more efficient than the LARS algorithm for large scale problems.
For a fair comparison with ESC, we compute an affinity graph for SSC using the same procedure as that used for ESC, i.e., the procedure in Algorithm~\ref{alg:esc}.

We also compare our method with $k$-means clustering (K-means) and spectral clustering on the $k$-nearest neighbors graph (Spectral).
It is known~\cite{Heckel:TIT15} that Spectral is a provably correct method for subspace clustering.
The $k$-means and $k$-d trees algorithms used to compute the $k$-nearest neighbor graph in Spectral are implemented using the VLFeat toolbox~\cite{vedaldi:vlfeat}. 
In addition, we compare with the three subspace clustering algorithms SSC-OMP \cite{You:CVPR16-SSCOMP}, OLRSC \cite{Shen:ICML16} and SBC \cite{Adler:TNNLS15}, which are able to handle large-scale data.
{
For experiments on the Extended Yale B database, we also include a comparison with LRR~\cite{Liu:ICML10} and $\ell_0$-SSC~\cite{Yang:ECCV16}, which cannot effectively handle EMNIST and GTSRB due to memory and running time constraints. 
For all subspace clustering methods (i.e., SSC-OMP, OLRSC, SBC, LRR and $\ell_0$-SSC) we use the code provided by their respective authors. 
}

To demonstrate the advantage of our exemplar selection method, we compare ESC-FFS to an approach we call ESC-Rand, which consists of selecting the exemplars $\cX_0$  at random from $\cX$, i.e., we replace the exemplar selection via FFS in step~\ref{step:esc-ffs} of Algorithm~\ref{alg:esc} by selecting $k$ atoms at random from $\cX$ to form $\cX_0$.
{
In experiments on Extended Yale B database, we further compare with methods where FFS in step~\ref{step:esc-ffs} of Algorithm~\ref{alg:esc} is replaced by other exemplar selection methods including $k$-centers, $K$-medoids~\cite{Park:ESA09}, SMRS~\cite{Elhamifar:CVPR12}, kDPP~\cite{Kulesza:ICML11}, two algorithms for separable NMF (i.e., SPA~\cite{Ren:TAES03,Gillis:TPAMI13} and Xray~\cite{Kumar:ICML13}), and two algorithms for column subset selection (i.e., GreedyCSS~\cite{Farahat:KIS15} and IPM~\cite{Joneidi:CVPR2019}).
For $k$-centers, we implement the farthest first traversal algorithm (see, e.g.~\cite{Williamson:Cambridge2011}).
For $K$-medoids, we use the function provided by \textregistered Matlab, which employs a variant of the algorithm in~\cite{Park:ESA09}.
For SMRS, kDPP and GreedyCSS, we use the code provided by their respective authors.
For SPA and Xray, we use the code provided by \cite{Gillis:SIAM15}. 
For IPM, we use our own implementation following the description in~\cite{Joneidi:CVPR2019}. 
}

\begin{figure*}[!t]
	\centering
	\subfloat[\label{fig:EMNIST_A} Accuracy]{\includegraphics[scale=0.65]{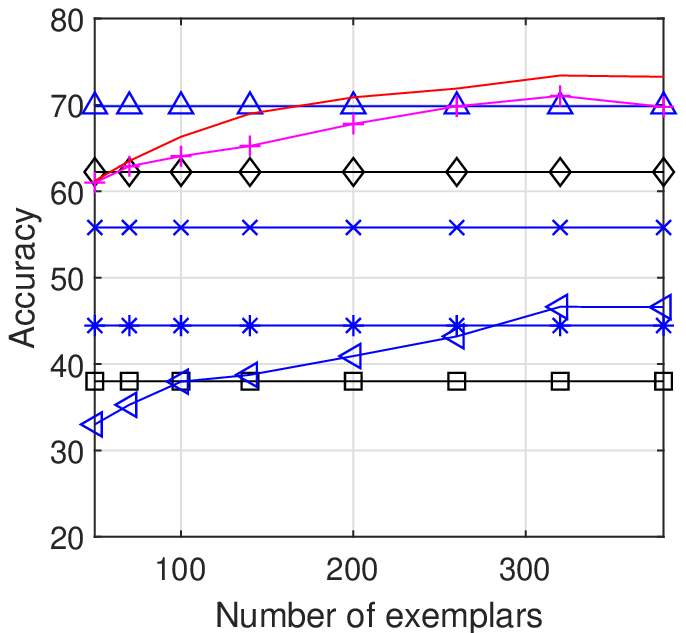}}
	~~
	\subfloat[\label{fig:EMNIST_F} F-score]{\includegraphics[scale=0.65]{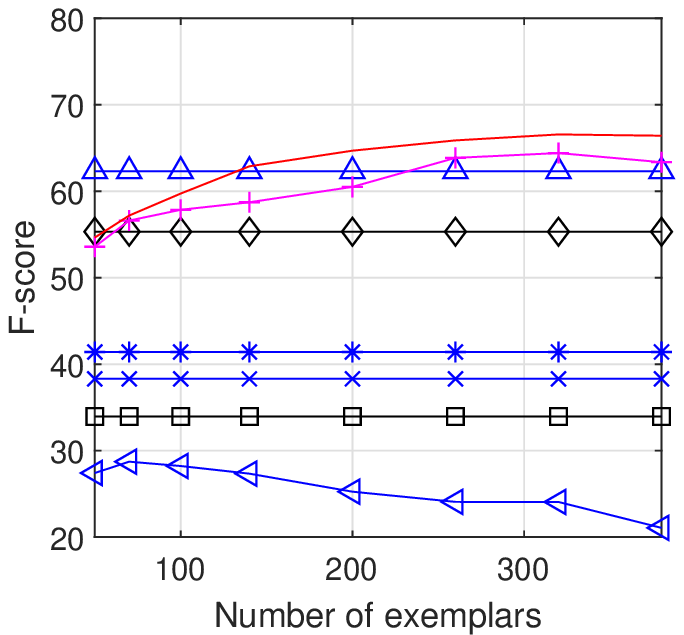}}
	~~
	\subfloat[\label{fig:EMNIST_T} Running time]{\includegraphics[scale=0.65]{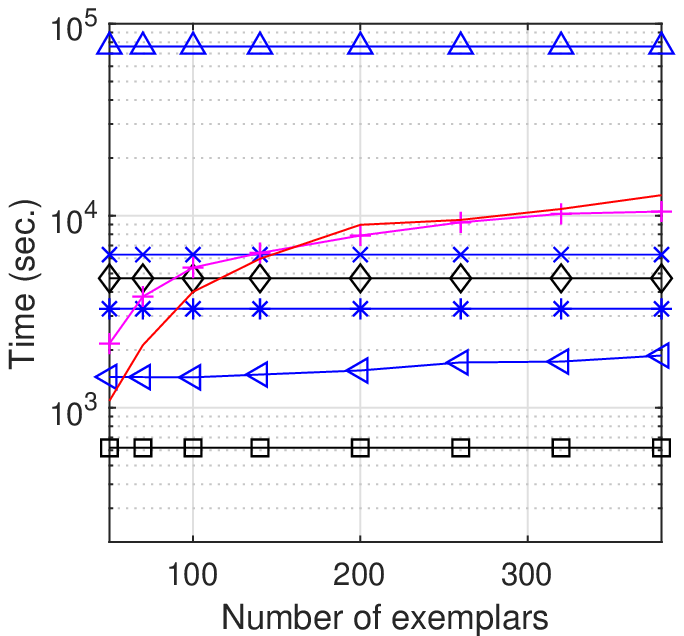}}
	\subfloat{\includegraphics[scale=0.6,trim={9.5cm 4cm 1cm 1cm},clip]{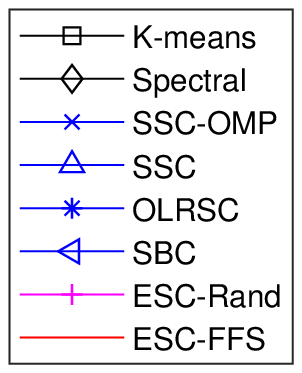}}
	\caption{\label{fig:EMNIST} Subspace clustering on $190,\!998$ images corresponding to $26$ lower case letters from the EMNIST database. We report the averaged accuracy, F-score and running time (in sec.) from $10$ trials. }
\end{figure*}

\myparagraph{Evaluation metrics} The first metric we use is the clustering accuracy.
It measures the maximum proportion of points that are correctly labeled over all possible permutations of the labels. 
Concretely, let $\{C_1, \cdots, C_n\}$ be the ground-truth partition of the data, $\{G_1, \cdots, G_n\}$ be a clustering result of the same data, $n_{ij} = |C_i \cap G_j|$ be the number of common objects in $C_i$ and $G_j$, and $\Pi$ be the set of all permutations of $\{1, \cdots, n\}$. The clustering accuracy is defined as
\begin{equation}
\text{Accuracy}=\max_{\pi\in\Pi}\frac{100}{N} \sum_{i=1}^n n_{i, \pi(i)}.
\end{equation}

In the context of classification, accuracy has been known to be biased when the dataset is class imbalanced \cite{Brodersen:ICPR10}. For example, if $99\%$ of a dataset consists of samples from one particular class, then assigning all data points to the same label yields at least $99\%$ accuracy. To address this issue, we also use the F-score averaged over all classes. Let $p_{ij} = n_{ij} / |G_j|$ be the precision and $r_{ij} = n_{ij} / |C_i|$ be the recall. The F-score between the clustering result $G_i$ and the  true class $C_j$ is defined as $F_{ij}=\frac{2p_{ij}r_{ij}}{p_{ij} + r_{ij}}$. We report the average F-score given by
\begin{equation}
\text{F-score}=\max_{\pi\in\Pi} \frac{100}{n} \sum_{i=1}^n F_{i, \pi(i)}.
\end{equation}


\myparagraph{Results on EMNIST}
Figure~\ref{fig:EMNIST} shows the results on EMNIST. From left to right, the sub-figures show, respectively, the accuracy, the F-score and the running time (Y axis) as a function of the number of exemplars (X axis). 
ESC-FFS outperforms all methods except SSC in terms of  accuracy and F-score when the number of exemplars is greater than~$70$.

Recall that in SSC each data point is expressed as a linear combination of all other points.
By selecting a subset of exemplars and expressing points using these exemplars, ESC-FFS is able to outperform SSC when the number of exemplars reaches $200$. 
In contrast, ESC-Rand does not outperform SSC by a significant amount, showing the importance of exemplar selection by FFS.


In terms of running time, we see that ESC-FFS is faster than SSC by a large margin.
Specifically, ESC-FFS is almost as efficient as ESC-Rand, which indicates that the proposed FFS Algorithm~\ref{alg:ffs-efficient} is efficient.

\begin{table}[!t]
	\centering
	\renewcommand{\arraystretch}{1.3}
	\caption{\label{tbl:gtsrb}Subspace Clustering on the GTSRB database. The parameter $k = 160$ is used for ESC-Rand and ESC-FFS. We report the mean and standard deviation for accuracy, F-score and running time (in sec.) from $10$ trials. }
	\begin{tabular}{c|ccc}
		Methods  &       Accuracy        &        F-score        &      Time (sec.)      \\ \hline
		$K$-means &     $63.7\pm3.5$      &     $54.4\pm2.8$      & $\mathbf{12.2}\pm0.5$ \\
		Spectral  &     $89.5\pm1.3$      &     $79.8\pm2.5$      &     $40.3\pm0.7$      \\ \hline
		SSC-OMP    &     $82.8\pm0.8$      &     $67.8\pm0.5$      &     $22.0\pm0.2$      \\
		SSC    &     $92.4\pm1.1$      &     $82.3\pm2.8$      &     $52.2\pm0.7$      \\
		OLRSC   &     $71.6\pm4.3$      &     $66.7\pm4.7$      &     $64.9\pm1.6$      \\
		SBC    &     $74.9\pm5.2$      &     $72.2\pm8.5$      &     $41.9\pm0.4$      \\ \hline
		ESC-Rand  &     $89.7\pm1.6$      &     $75.5\pm4.9$      &     $21.5\pm0.4$      \\
		ESC-FFS (ours)  & $\mathbf{93.0}\pm1.3$ & $\mathbf{85.3}\pm2.5$ &     $25.2\pm1.2$      \\ \hline
	\end{tabular}
\end{table}

\myparagraph{Results on GTSRB}
Table~\ref{tbl:gtsrb} reports the clustering performance on the GTSRB database. In addition to reporting average performances, we report the standard deviations. 
The variation in accuracy and F-score across trials is due to 1) random initializations of the $k$-means algorithm, which is used (trivially) in the K-means method, and in the spectral clustering step of all other methods, and 2) random dictionary initialization in OLRSC, SBC, ESC-Rand and ESC-FFS. 

We observe that ESC-FFS outperforms all the other methods in terms of accuracy and F-score. In particular, ESC-FFS outperforms SSC, which in turn outperforms ESC-Rand, thus showing the importance of finding a good representative set of exemplars and the effectiveness of FFS in achieving this.
In addition, the standard deviation of the accuracy and F-score values for ESC-Rand are larger than for ESC-FFS. 
This indicates that the set of exemplars given by FFS is more robust in giving reliable clustering results than the randomly selected exemplars in ESC-Rand.
In terms of running time, ESC-FFS is also competitive.

{
\myparagraph{Results on Extended Yale B}
The averaged accuracy, F-score and running time over $10$ randomly sampled imbalanced subsets of the Extended Yale B database are reported in Table~\ref{tbl:eyaleb_clustering}.
Observe that the $\ell_0$-SSC and LRR have slightly better performance than SSC, but still are not able to effectively handle imbalanced data. 
On the other hand, the ESC methods with exemplar selection by $k$DPP, SMRS, SPA, Xray, GreedyCSS and FFS all have higher accuracies and F-scores than SSC, demonstrating the effectiveness of the exemplar selection approach for handling imbalanced data. 
In particular, the FFS produces the highest accuracy and the second highest F-score. 
The $k$-centers and $K$-medoids do not demonstrate a significant gain from ESC-Rand. 
This is because images of the same face lie approximately in a subspace, and their pairwise distances may not be small.

\begin{table}[t]
	\centering
	\caption{Subspace Clustering on the Extended Yale B database. The parameter $k = 250$ is used for ESC-FFS. We report the mean and standard deviation for accuracy, F-score and running time (in sec.) from $10$ trials. }
	\renewcommand{\arraystretch}{1.3}
	\label{tbl:eyaleb_clustering}
	\begin{tabular}{c|ccc}
		Methods     &       Accuracy        &        F-score        &     Time (sec.)      \\
		\hline
		$K$-means    &     $20.1\pm4.0$      &    $18.9 \pm 3.9$     & $\mathbf{1.0\pm0.2}$ \\
		Spectral     &     $46.5\pm4.5$      &    $44.0 \pm 5.0$     & $\mathbf{0.2\pm1.4}$ \\
		\hline
		SSC-OMP     &     $56.8\pm 4.2$     &    $49.9 \pm 2.4$     &    $0.4 \pm 0.03$    \\
		SSC       &     $67.1\pm3.6$      &     $60.3\pm4.5$      &     $4.6\pm0.2$      \\
		OLRSC      &     $30.7\pm3.1$      &     $29.3\pm 3.6$     &     $1.9\pm0.2$      \\
		SBC       &     $45.4\pm5.6$      &     $43.6\pm6.2$      &     $4.0\pm0.1$      \\
		$\ell_0$-SSC   &     $67.2\pm3.6$      &     $60.4\pm4.5$      &     $4.6\pm0.3$      \\
		LRR       &     $68.3\pm5.3$      &     $61.7\pm5.4$      &     $12.3\pm0.7$     \\
		\hline
		ESC-Rand     &     $65.7\pm6.3$      &     $59.6\pm8.1$      &     $1.4\pm0.04$     \\
		ESC-$k$-centers &     $67.0\pm4.0$      &     $58.9\pm3.9$      &     $2.0\pm0.1$      \\
		ESC-$K$-medoids &     $64.5\pm4.7$      &     $56.9\pm5.6$      &     $5.7\pm0.2$      \\
		ESC-$k$DPP    &     $69.7\pm5.7$      & $\mathbf{63.0\pm6.9}$ &     $8.1\pm0.7$      \\
		ESC-SMRS     &     $67.9\pm5.2$      &     $60.7\pm6.1$      &     $11.2\pm2.2$     \\
		ESC-SPA     &     $67.2\pm5.2$      &     $60.0\pm4.9$      &     $1.8\pm0.2$      \\
		ESC-Xray     &     $67.8\pm5.3$      &     $60.5\pm4.6$      &    $231.5\pm18.5$    \\
		ESC-GreedyCSS  & $\mathbf{70.3\pm3.3}$ &     $61.4\pm2.9$      &     $1.5\pm0.1$      \\
		ESC-IPM     &     $63.4\pm4.2$      &     $56.9\pm4.4$      &     $11.2\pm2.2$     \\
		ESC-FFS (ours)  & $\mathbf{71.1\pm4.6}$ & $\mathbf{62.4\pm6.3}$ &     $3.2\pm0.3$      \\
		\hline
	\end{tabular}
\end{table}

\begin{table}[t]
	\centering
	\caption{Effect of varying the parameter $t$ for subspace clustering on Extended Yale B database using ESC-FFS. }
	\renewcommand{\arraystretch}{1.3}
	\label{tbl:eyaleb_clustering_t}
	\begin{tabular}{c|cccccc}
		t     &  $2$   &  $3$   &  $4$   &  $5$   &  $6$   & $8$ \\
		\hline\hline
		Accuracy & $61.2$ & $70.9$ & $68.3$ & $67.5$ & $65.7$ & $60.4$    \\\hline
		F-score  & $54.4$ & $62.1$ & $63.5$ & $62.5$ & $62.5$ & $57.1$
	\end{tabular}
\end{table}

\begin{table*}[!t]
	\centering
	\caption{\label{tbl:eyaleb-subset-selection} Classification from subsets on the Extended Yale B face database. We report the mean and standard deviation for classification accuracy (\%) and running time of the subset selection from 50 trials.}
	\renewcommand{\arraystretch}{1.3}
	
	\begin{tabular}{c|ccc|c}
		Methods         &          NN           &          SRC          &          SVM          &   Time (sec.)    \\
		\hline
		Rand          &     $72.0\pm2.8$      &     $85.4\pm2.1$      &     $84.4\pm2.4$      & $\mathbf{<1e-3}$ \\
		$k$-centers       &     $72.8\pm3.5$      &     $86.0\pm2.4$      &     $84.1\pm2.8$      &   $0.2\pm0.01$   \\
		$K$-medoids       &     $78.2\pm2.7$      &     $87.0\pm1.9$      &     $86.7\pm2.0$      &   $1.6\pm0.1$    \\
		$k$DPP         &     $72.8\pm3.3$      &     $88.5\pm1.8$      &     $88.7\pm2.3$      &   $0.4\pm0.01$   \\
		SMRS          &     $72.0\pm2.9$      &     $83.8\pm2.4$      &     $82.8\pm2.7$      &   $3.1\pm0.2$    \\
		{SPA}    &     $68.8\pm4.6$      &     $89.1\pm4.2$      &     $90.1\pm4.3$      &   $0.1\pm0.1$    \\
		{Xray}    &     $65.0\pm6.7$      &     $84.3\pm7.1$      &     $83.9\pm8.4$      &   $29.0\pm2.9$   \\
		{IPM}    &     $66.4\pm3.4$      &     $83.4\pm3.0$      &     $83.2\pm2.8$      &   $2.2\pm0.3$    \\
		{GreedyCSS} & $\mathbf{79.4\pm2.8}$ & $\mathbf{92.1\pm1.3}$ &     $91.8\pm1.7$      &  $0.04\pm0.004$  \\
		\hline
		FFS  (ours)       &     $70.0\pm3.4$      & $\mathbf{92.1\pm2.1}$ & $\mathbf{91.9}\pm2.5$ &   $0.7\pm0.1$    \\
		\hline
	\end{tabular}
	
\end{table*}

\myparagraph{Effect of parameter $t$}
In Algorithm~\ref{alg:esc}, segmentation of the data $\cX$ from the selected exemplars $\cX_0$ is computed by finding the $t$-nearest neighbors of the representation vector for each data point, then applying spectral clustering to the $t$NN graph. 
To understand the role of the parameter $t$, we conduct additional experiments on the Extended Yale B dataset and report clustering accuracy with varying values of $t$ in Table~\ref{tbl:eyaleb_clustering_t}. 
We can see that both the Accuracy and the F-score are relatively stable for $t$ in the range $[3, 6]$.
In addition, we explore an alternative method for computing the segmentation from exemplars. 
In particular, we replace step~\ref{step:knn} and step~\ref{step:affinity} of Algorithm~\ref{alg:esc} with the $k$-means algorithm applied to the set of normalized representation vectors $\{\c_j / \|\c_j\|_2\}$. 
This produces an accuracy of $41.6\%$ and an F-score of $40.4\%$, which are much lower than those obtained with $t$-nearest neighbors based approaches. 
}

\subsection{Exemplar selection for classification}

In this section, we evaluate the performance of the FFS algorithm as a tool for selecting a subset of representatives that is subsequently used to classify the entire data set as described in Algorithm~\ref{alg:es4c}. 
The parameter $\lambda$ in Algorithm~\ref{alg:es4c} is set to $200$.  
The evaluation is performed with the randomly sampled imbalanced subsets of the Extended Yale B database as described in Section~\ref{sec:esc-experiments-subspace-clustering}.

In particular, we apply each exemplar selection method to select $100$ images from the dataset.
Note that during this phase we assume that the ground truth labeling is unknown.
After the exemplars have been selected, we assume that the labels of the exemplars are given and use the sparse representation based classification (SRC) method described in Algorithm~\ref{alg:es4c} to assign a label to each of the rest of the data point. 
In addition to SRC, we also report the results given by the nearest neighbor (NN) and the linear support vector machine (SVM) classifiers.

In Table~\ref{tbl:eyaleb-subset-selection} we report the classification accuracy averaged over $50$ trials. 
We see that the performance with the SRC and SVM classifiers is significantly better than with the NN classifier, which is due to the fact that images of the same face lie approximately in a subspace and their pairwise distance is not necessarily small. 
In particular, our method obtains the highest accuracy with SRC and SVM. 


\begin{figure}[!t]
	\centering
	\subfloat[\label{fig:EYaleB_classify_lambda_knn} NN]{\includegraphics[scale=0.65]{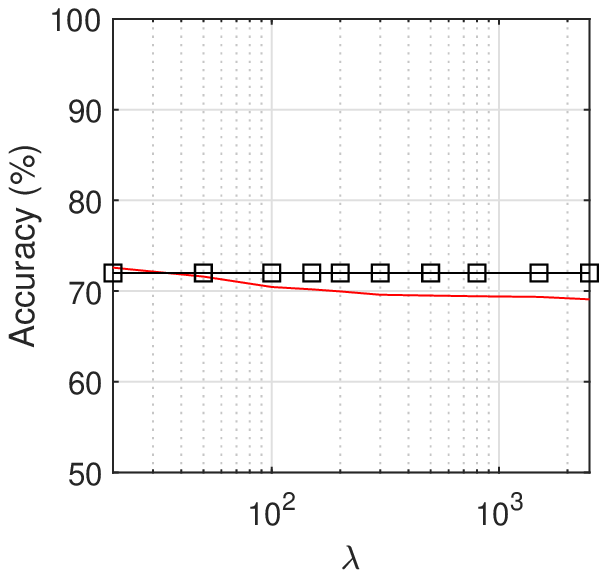}}
	~
	\subfloat[\label{fig:EYaleB_classify_lambda_src} SRC]{\includegraphics[scale=0.65]{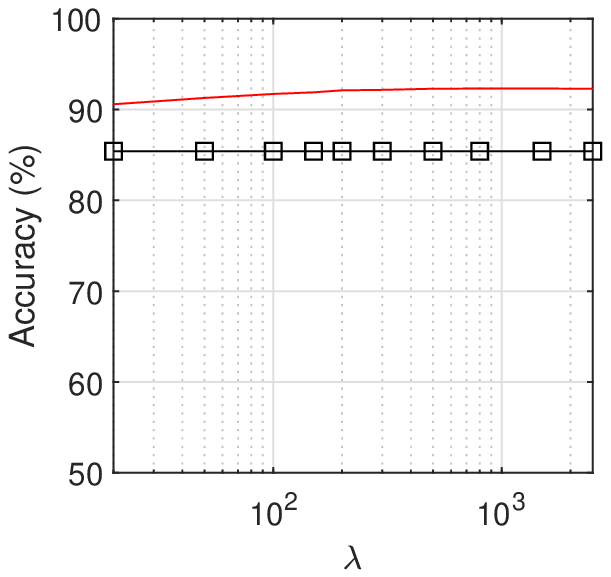}}
		\\
	\subfloat[\label{fig:EYaleB_classify_lambda_svm} SVM]{\includegraphics[scale=0.65]{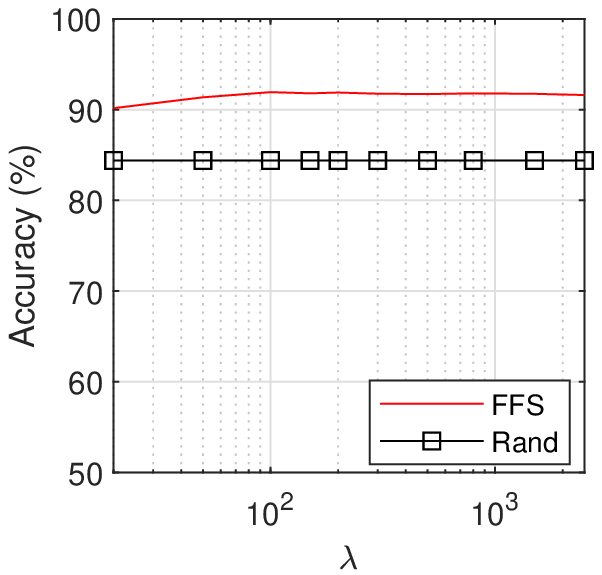}}
		~
	\subfloat[\label{fig:EYaleB_classify_lambda_t} Time]{\includegraphics[scale=0.65]{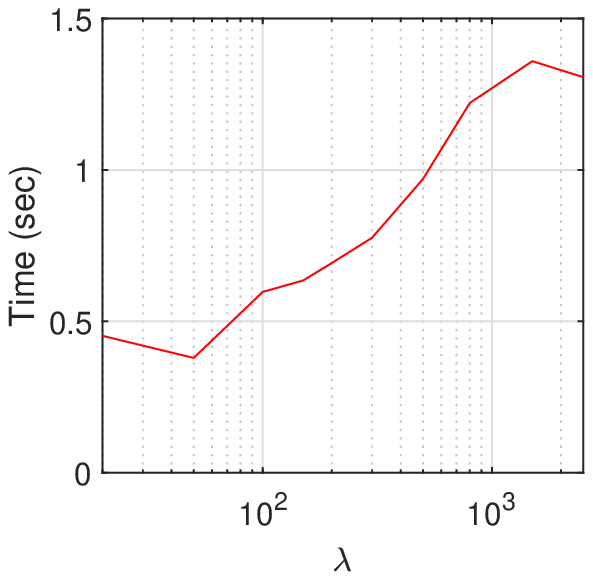}}
	\caption{\label{fig:EYaleB_classify_lambda} Effect of varying the parameter $\lambda$ for classification on Extended Yale B database. The parameter $\lambda$ is varied along the $x$-axis from $20$ to $2500$. Note that the $x$-axis is in log scale. }
\end{figure}

\begin{figure}[!htb]
	\centering
	\includegraphics[scale=0.8]{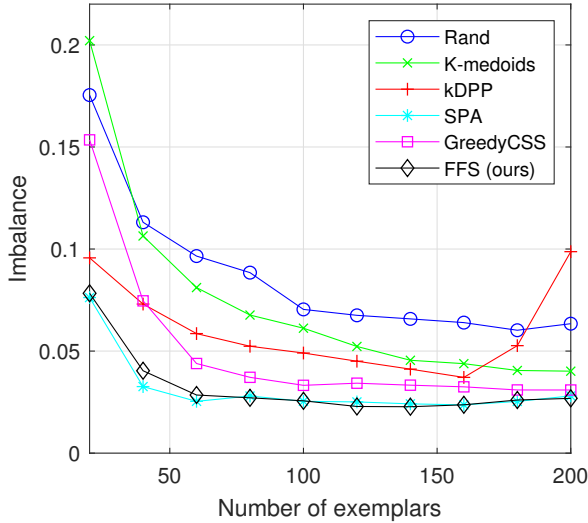}
	\caption{\label{fig:balance} Performance of exemplar selection for finding a balanced set of representatives from imbalanced classes in the Extended Yale B dataset. We test the methods with the number of representatives varied in the $x$-axis from $20$ to $200$, and report the averaged imbalancedness measure from $10$ trials.  }
\end{figure}

\myparagraph{Effect of parameter $\lambda$} {
To understand the effect of the parameter $\lambda$ in our FFS algorithm, we conduct experiments with varying values of $\lambda$ in the range of $[20, 2500]$ and report the classification accuracy and running time in Figure~\ref{fig:EYaleB_classify_lambda}. 
For comparison purposes, we also plot the curves for Rand. 
It can be seen from Figure~\ref{fig:EYaleB_classify_lambda_src} and Figure~\ref{fig:EYaleB_classify_lambda_svm} that the accuracy of FFS with SRC and SVM classifiers is non-decreasing as a function of $\lambda$ in the range of $[20, 200]$ and is mostly insensitive to $\lambda$ in the range $\lambda>200$. 
On the other hand, Figure~\ref{fig:EYaleB_classify_lambda_t} shows that the running time of FFS increases significantly with $\lambda$. 
This is because the LASSO version of the LARS algorithm that we adopt for solving the optimization problem \eqref{eq:representation-f} computes the entire regularization path, therefore a larger value of $\lambda$ requires more computation. 
}

\myparagraph{Measuring the imbalance} {
We further evaluate the ability of FFS to handle imbalanced data by measuring the degree of imbalance for the selected representatives. 
Specifically, since the $64$ images in each class of the Extended Yale B database are captured with $64$ strobes mounted on a fixed illumination rig, we may assume that all the classes have the same within-class variation and that a best set of exemplars contains equal number of samples from each of the $10$ classes. 
To quantitatively measure the degree of imbalance, we compute the entropy of the proportion of exemplars that are selected from each class. 
That is, we compute
\begin{equation}
	\text{Entropy} = -\sum_{i=1}^{10} p_i \log p_i, ~~\text{where}~p_i = \frac{s_i}{\sum_{j=1}^{10}s_j},
\end{equation}
and $s_i$ is the number of exemplars selected from the $i$-th class. 
The entropy is equal to one if and only if $s_1 = s_2 = \cdots = s_{10}$ and is zero if no data point is selected from at least one class. 
Then, we define the imbalance as one minus the entropy, i.e.,
\begin{equation}
\text{Imbalance} = 1 - \text{Entropy}.
\end{equation}
Therefore, the imbalance is a non-negative number that takes value $0$ if and only if the set of exemplars contains equal number of data points from each class. 

The results for FFS, Rand and four other methods that obtain the highest performance in the classification task (i.e., $K$-medoids, $k$DPP, SPA and GreedyCSS) are reported in Figure~\ref{fig:balance}.
We can see that all methods produce lower imbalance as the number of exemplars increases in the range $[20, 160]$. 
Note that among the $10$ classes in our dataset, each of the $3$ smallest classes has only $16$ samples. 
Therefore, it is impossible for the selected set of exemplars to be balanced when the number of such exemplars is greater than $160$. 
This may explain why the imbalance for $k$DPP, SPA and FFS increases as the number of exemplars goes beyond $160$. 

In comparing different methods, we observe that random sampling
fails to produce a balanced subset because the original data
is highly imbalanced. 
Our FFS significantly outperforms all the other methods for all sizes of subset, with the only exception being SPA, which performs on par with our FFS. 

}

\section{Conclusion}

We presented a novel approach for unsupervised exemplar selection in a union of subspaces. 
Our method searches for a set of exemplars from the given dataset such that all data points can be well-represented by the exemplars in terms of a sparse representation cost. 
When the data comes from a union of subspaces, we proved that our method selects a set of exemplars that is able to represent all data points. 
We also introduced an algorithm for approximately solving the exemplar selection optimization problem. 
Empirically, we demonstrated that the exemplars selected by our method can be used for generating a segmentation of the dataset.

\appendix[Relation to the sphere covering problem]

We consider the special case when the dataset $\cX$ coincides with the unit sphere of $\Re^D$, i.e., $\cX = \Sp^{D-1}$. 
In this case, we establish that our exemplar selection objective in \eqref{eq:exemplar-objective} is related to finding $k$ points on the unit sphere with minimal \emph{covering radius}, 
which is defined in the following.
\begin{definition}[Covering radius]\label{def:covering-radius}
	The covering radius of a set of points $\cV \subseteq \Sp^{D-1}$ is defined as
	\begin{equation}
	\gamma(\cV) := \max_{\w \in \Sp^{D-1}} \min_{\v \in \cV} ~ \cos ^{-1} ( \langle \v, \w \rangle) .
	\end{equation}
\end{definition}
\noindent
The covering radius of the set $\cV$ is the minimum angle such that the union of spherical caps centered at each point in $\cV$ with this angle covers the entire unit sphere $\Sp^{D-1}$. Our next result establishes a relationship between the covering radius and our cost function. The proof of the result uses the inradius of a convex body, which is defined as follows.
\begin{definition}[inradius]\label{def:inradius}
		The inradius of a convex set $\cK$, denoted by $r(\cK)$, is the radius of the largest Euclidean ball inscribed in $\cK$.
\end{definition}

\begin{lemma}
	For any finite $\cX_0 \subseteq \cX = \Sp^{D-1}$, it holds that $F_\infty(\cX_0) = 1 / \cos \gamma(\pm \cX_0)$.
	\label{thm:cost-coveringradius}
\end{lemma}
\begin{proof}
	From Definition \ref{def:representation-cost}, \eqref{eq:basis-pursuit-Minkowski}, and Definition \ref{def:Minkowski} we have 
	\begin{align}
	F_\infty(\cX_0) 
	&= \sup_{\x \in \Sp^{D-1}} f_\infty(\x, \cX_0) = \sup_{\x \in \Sp^{D-1}} \|\x\|_{\cK_0} \nonumber \\
	&= \sup_{\x \in \Sp^{D-1}} \inf\{t>0: \x /t \in \cK_0\}.
	\label{eq:prf-F-r-pre}
	\end{align}
	Then, using Definition~\ref{def:inradius} and the symmetry of $\cK_0$, we have
	\begin{equation}
	\begin{split}
	r(\cK_0) &= \sup\{r > 0: r \x \in \cK_0 \ \text{for all} \ \x \in \Sp^{D-1}\}\\
	&= \inf_{\x \in \Sp^{D-1}}\sup\{r > 0: r \x \in \cK_0\}\\
	&= \inf_{\x \in \Sp^{D-1}}\frac{1}{\inf\{t > 0: \x/t \in \cK_0\}}\\
	&= \frac{1}{\sup_{\x \in \Sp^{D-1}}\inf\{t > 0: \x/t \in \cK_0\}}.
	\label{eq:inradius}
	\end{split}
	\end{equation}
	By comparing \eqref{eq:prf-F-r-pre} and \eqref{eq:inradius} we have 
	\begin{equation}
	F_\infty(\cX_0) = 1 / r(\cK_0).
	\label{eq:prf-F-r}
	\end{equation}
	Furthermore, it was shown in \cite[Theorem 9]{You:18} that
	$r(\cK_0) = \cos \gamma(\pm \cX_0)$.
	Combining this result with \eqref{eq:prf-F-r} allows us to conclude that $F_\infty(\cX_0) = 1/\cos \gamma(\pm \cX_0)$, as claimed.
\end{proof}

It follows from Lemma~\ref{thm:cost-coveringradius} that $\arg\min_{|\cX_0|\le k} F_\infty(\cX_0)= \arg\min_{|\cX_0|\le k} \gamma(\pm \cX_0)$ when $\cX = \Sp^{D-1}$, i.e., the exemplars $\cX_0$ selected by \eqref{eq:exemplar-objective} constitute a solution to the problem of finding a subset of $\cX = \Sp^{D-1}$ of size $k$ with minimum covering radius. Note that the covering radius $\gamma(\pm \cX_0)$ of the subset $\cX_0$ with $|\cX_0| \le k$ is minimized when the points in the symmetrized set $\pm \cX_0$ are uniformly distributed on  $\Sp^{D-1}$. The problem of equally distributing points on the sphere without symmetrizing them, i.e. $\min_{|\cX_0| \le k} \gamma(\cX_0)$, is known as the sphere covering problem.
This problem was first studied by \cite{Toth:1943} and remains unsolved in geometry~\cite{Croft:1991}.


\ifCLASSOPTIONcompsoc
  \section*{Acknowledgments}
\else
  \section*{Acknowledgment}
\fi

C. You, D. P. Robinson and R. Vidal are supported by NSF grant 1618637. C. Li is supported by IARPA grant 127228.

\ifCLASSOPTIONcaptionsoff
  \newpage
\fi



\bibliographystyle{IEEEtran}
\bibliography{biblio/vidal,biblio/vision,biblio/math,biblio/learning,biblio/sparse,biblio/geometry,biblio/dti,biblio/recognition,biblio/surgery,biblio/coding,biblio/segmentation,biblio/dataset,biblio/collaborator/robinson}
\end{document}